\documentclass[acmsmall]{acmart}
\usepackage{graphicx}
\usepackage{subfigure}
\usepackage{multirow}
\usepackage{bm}
\usepackage{float}
\usepackage[linesnumbered,commentsnumbered,ruled,vlined]{algorithm2e}

\AtBeginDocument{%
  \providecommand\BibTeX{{%
    \normalfont B\kern-0.5em{\scshape i\kern-0.25em b}\kern-0.8em\TeX}}}

\setcopyright{acmcopyright}
\copyrightyear{2018}
\acmYear{2018}
\acmDOI{10.1145/1122445.1122456}

\acmJournal{JACM}
\acmVolume{37}
\acmNumber{4}
\acmArticle{111}
\acmMonth{8}

\begin{document}

%\title{Reimagining City Configuration: Automated Urban Planning via Adversarial Learning}

\title{Automated Urban Planning for Reimagining City Configuration via  Adversarial Learning: Quantification, Generation, and Evaluation}

% \affiliation{%

\author{Dongjie Wang}
\affiliation{%
  \institution{University of Central Florida}
  \streetaddress{4328 Scorpius Street}
  \city{Orlando}
%   \state{FL}
  \country{USA}
  \postcode{FL 32816-2362}
  }
\email{wangdongjie@knights.ucf.edu}

\author{Yanjie Fu}
\affiliation{%
  \institution{University of Central Florida}
  \streetaddress{4328 Scorpius Street}
  \city{Orlando}
%   \state{FL}
  \country{USA}
  \postcode{FL 32816-2362}
}
\email{yanjie.fu@ucf.edu}

\author{Kunpeng Liu}
\affiliation{%
\institution{University of Central Florida}
  \streetaddress{4328 Scorpius Street}
  \city{Orlando}
%   \state{FL}
  \country{USA}
  \postcode{FL 32816-2362}
  }
 \email{kunpengliu@knights.ucf.edu}

\author{Fanglan Chen}
\affiliation{%
  \institution{Virginia Tech}
  \streetaddress{7054 Haycock Rd}
  \city{Falls Church}
%   \state{VA}
  \country{USA}}
  \postcode{VA 22043}
  \email{fanglanc@vt.edu}

\author{Pengyang Wang}
\affiliation{%
\institution{University of Macau}
  \streetaddress{Taipa}
  \city{Macau}
%   \state{FL}
  \country{China}
%   \postcode{FL 32816-2362}
  }
\email{pywang@um.edu.mo}

\author{Chang-Tien Lu}
\affiliation{%
  \institution{Virginia Tech}
  \streetaddress{7054 Haycock Rd}
  \city{Falls Church}
  \country{USA}
  \postcode{VA 22043}
  }
\email{ctlu@vt.edu}

% \affiliation{%

\renewcommand{\shortauthors}{Wang, et al.}

% Effective urban planning can help to mitigate the operational and social vulnerability of an urban system, such as high taxes, crimes, traffic accidents, and etc. 

\begin{abstract}
    Urban planning refers to the efforts of designing land-use configurations given a region. However, to obtain effective urban plans, urban experts have to spend much time and effort analyzing sophisticated planning constraints based on domain knowledge and personal experiences.
    To alleviate the heavy burden of them and produce consistent urban plans, we want to ask that can AI accelerate the urban planning process, so that human planners only adjust generated configurations for specific needs?
    The recent advance of deep generative models provides a possible answer, which inspires us to automate urban planning from an adversarial learning perspective. However, three major challenges arise: 1) how to define a quantitative land-use configuration? 2) how to automate configuration planning? 3) how to evaluate the quality of a generated configuration? In this paper, we systematically address the three challenges. Specifically, 1) We define a land-use configuration as a longitude-latitude-channel tensor. 2) We formulate the automated urban planning problem into a task of deep generative learning. The objective is to generate a configuration tensor given the surrounding contexts of a target region. In particular, we first construct spatial graphs using geographic and human mobility data crawled from websites to learn graph representations. We then combine each target area and its surrounding context representations as a tuple, and categorize all tuples into positive (well-planned areas) and negative samples (poorly-planned areas). Next, we develop an adversarial learning framework, in which a generator takes the surrounding context representations as input to generate a land-use configuration, and a discriminator learns to distinguish between positive and negative samples. 3) We provide quantitative evaluation metrics and conduct extensive experiments to demonstrate the effectiveness of our framework.
\end{abstract}



\maketitle

\section{Introduction}
%Urban planning is an essential part for a city, which determines the future development direction of the city. Effective urban planning is helpful to mitigate the vulnerability of a urban system, such as high crime rate, traffic congestion, air pollution, etc. The designing process of urban planning is interdisciplinary and complex, which involves many constraints such as public policy, social science, economics, and so on. In this paper, for better modeling and generation, we simplify urban planning as a land-use configuration generation task based on the surrounding spatial contexts.

We study the problem of machine learning for automated urban planning.
Urban planning is an interdisciplinary and complex process that involves public policy, social science, engineering, architecture, landscape, and other related fields. 
In this paper, we refer urban planning to the efforts of designing land-use configurations of a target region, which is a reduced yet essential task of urban planning ~\cite{van2013co}.
Effective urban planning can help to mitigate the operational and social vulnerability of a urban system, such as high tax, crimes, traffic congestion and accidents, pollution, depression, and anxiety ~\cite{yiftachel1989towards}.

\begin{figure}[htbp]
% \vspace{-0.3cm}
    \centering
    \includegraphics[width=0.9\textwidth]{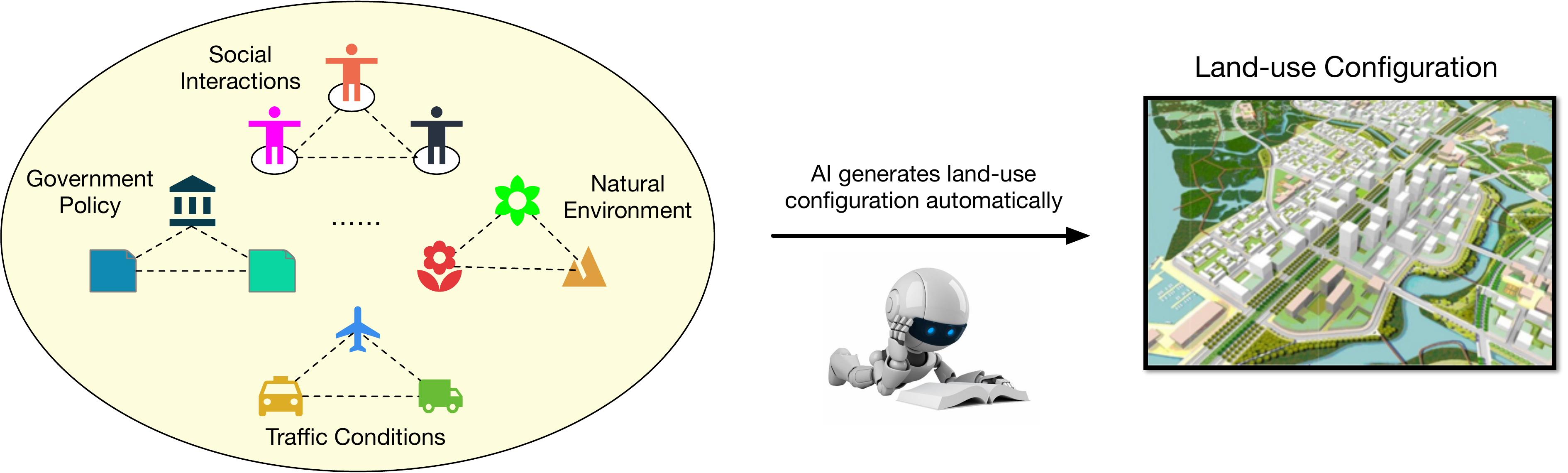}
    \caption{Our expectation is to build an AI model to generate the land-use configuration automatically by considering the constraints that are used in traditional urban planning process.}
    \label{residential_context}
    \vspace{-0.3cm}
\end{figure}

This observation motivates us to rethink urban planning in the era of artificial intelligence: What roles do deep learning play in urban planning? Can machines develop and learn at a human capability to automatically and quickly calculate land-use configurations? In this way, machines can be planning assistants and urban planning professionals can finally adjust machine-generated plans for specific needs.

Due to the high complexity and specificity of urban systems, urban planners need to consider and balance different various planning requirements, such as  proximity metrics (e.g., distances to important places), access indexes (e.g., accessibility to food, recreation, goods, services, entertainment,  transit, municipal services, mobility indexes), mobility indices (e.g., sidewalks, bike lanes, speed limits, crash rates), emergency responses (e.g., hospitals, fire departments) and, thus, planing highly replies on empirical experience and domain knowledge ~\cite{maheshwari2016balanced}.
As a result, it is highly appealing to pursuit a fast, automated, and machine-assisted planning strategy. 
The recent advance of deep learning, particularly deep adversarial and generative learning,  provide a great potential to teach a machine at a human capability  to design and generate city configurations ~\cite{zhang2020curb,zhang2019trafficgan,ding2018semi,liu2020catgan}. 
This inspiration motivates us to rethink urban planning from the lens of deep learning: can AI automate  the calculation of land-use configuration and the balancing of various planning factors, so professional planners can finally adjust machine-generated plans for specific needs? 

%With the great success of artificial intelligence in recent years, it motivates us to ask that can we use artificial intelligence to accelerate urban planning for reducing the heavy burdens of planners? The recent advance of deep learning, especially deep generative adversarial networks, provides a great potential for teaching a machine to imagine and create. Thus, it leads us to think that can we teach a machine to learn the designing capability of planners for calculating the land-use configuration solution quickly and automatically? If the answer is yes, the machine can be a planning assistant and human planners only need to adjust the machine-generated plans for specific needs.

All of the above evidences prompt us to develop a data-driven AI-enabled automated urban planner.
However, three unique challenges arise to achieve the goal:
(1) How can we quantify a land-use configuration plan? 
(2) How can we develop a deep adversarial generative learning framework to  learn the good and the bad of existing urban communities as data-driven knowledge, and, moreover, generate quality urban configuration?
(3) How can we evaluate the quality of generated land-use configurations?
Next, we will introduce our research insights and solutions for the three challenges.

First, as the objective is to teach a machine to generate the land-use configuration of a target region, it is naturally critical to define a machine- perceivable structure for the land-use configuration.
In practice, the land-use configuration plan of a region can be geographically defined by a set of Point of Interests (POIs) and their corresponding locations (e.g., latitudes and longitudes) and urban functionality categories (e.g., shopping, banks, education, entertainment, residential).
A close look can reveal that a land-use configuration is  a high-dimensional indicator that precisely illustrates what, where, and how many we should build in a target region.
After exploring large-scale land-use data, we observe that there is not just location-location statistical autocorrelation but also location-functionality statistical autocorrelation in a land-use configuration. To preserve such statistical correlations, we propose to represent a land-use configuration  as a latitude-longitude-channel tensor, where each channel is a specific category of POIs that are distributed across the target area, and the value of an entry in the tensor is the number of POIs. In this way, the tensor not just describes the location-location interaction, but also captures location-function interaction.

Second, after we quantitatively define the land-use configuration, the next question is that how to teach a machine to automatically generate a land-use configuration?
We analyze large-scale urban residential community data, and find that:
(1) an urban community can be viewed as an attributed node in a socioeconomic network (city as a community-community network), and this node proactively interacts with surrounding nodes (environments);
(2) the coupling, interaction, and coordination of a community and surrounding environments significantly influence the livability, vibrancy, and quality of a community. 
Based on the above observations, we aim to develop a function that map the surrounding contexts to  a well-planned configuration tensor. 
Recently, the development of deep generative and adversarial learning provides a great potential for solving this problem.
We reformulate the task into an adversarial learning paradigm, in which: 
(1) A neural generator is analogized as a machine planner that generates a land-use configuration; 
(2) The generator generates a configuration in terms of the feature representation of surrounding spatial contexts; 
(3) The surrounding context feature representation is learned via self-supervised representation learning collectively from spatial graphs.
(4) A neural discriminator is to classify whether the generated land-use configuration is well-planned (positive) or poorly-planned (negative). 
(5) A new mini-max loss function is constructed to guide the generator to learn  the configuration patterns of well-planned areas, compared to poorly-planned areas.

Third, how can we evaluate the quality of a generated land-use configuration? This has been a long-standing challenging question.
The most solid and sound validation is to collaborate with urban developers and city agencies to implement a machine-generated configuration into a target region to observe the development of the region in the following years. 
However, the validation method is not practical in reality. 
In this paper, we design and develop three strategies to assess the generated configurations: 
(1) We leverage different distance measurements to measure the similarity between  generated configurations and well-planned configurations.
If the distance is small, it indicates that our generated configurations preserve the overarching distribution characteristics of well-planned configurations.
(2) We develop a scoring model to score the quality of the generated configurations.
Specifically, since we have collected a set of existing land-use configurations and 0-1 labels (1: well-planned 0: poorly-planned) as training data, we train a regression model to predict the quality score ranging from 0 to 1.
After that, given a machine-generated configuration as testing data, we use the regression model to predict its corresponding score. 
(3) We use a variety of visualization approaches to visualize the generated configurations, so domain experts can evaluate the generated quality and rationality.

Our preliminary work in ~\cite{wang2020reimagining} proposed a fundamental automated urban planning framework to automatically generate land-use configurations. 
The preliminary framework can be further improved to enhance its stability and efficiency from a computational perspective. 
For this purpose, in this journal version, we develop a new conditioning augmentation module adding to the preliminary framework to enhance its performance.
Specifically, we first estimate the distribution of the embedding space of surrounding spatial contexts.
Then, we sample embedding vectors from the estimated distribution of the embedding space to replace the embeddings of surrounding spatial contexts. 
The benefit is to use embedding space distribution estimation to augment data and overcome the sparsity of  surrounding spatial context data. 
Later, we propose a new loss function that considers the embedding space regularization and standardization of surrounding spatial contexts.
The new loss function can accelerate the convergence and improve the efficiency of learning. 
In addition, aside from prediction-based and visualization-based valiation approaches~\cite{wang2020reimagining}, in this paper, we design a distance-based strategy to evaluate the quality of machine-generated configuration plans. 

In summary, in both our preliminary work~\cite{wang2020reimagining} and this extended version, we develop an adversarial learning framework to generate effective land-use configurations by learning from urban geography, human mobility, and socioeconomic data. 
Specifically, our contributions are:
1) We develop a latitude-longitude-channel tensor to quantify a land-use configuration plan. 
2) We propose a socioeconomic interaction perspective to understand urban planning as a process of optimizing the coupling between a community and surrounding environments. 
3) We reformulate the automated urban planning problem into an adversarial learning framework that maps surrounding spatial contexts into a configuration tensor. 
4) We computationally enhance the efficiency and stability of the proposed framework by devising a conditioning augmentation module via leveraging a new sampling technique and a new optimization loss function. 
5) We develop multiple strategies (i.e., distance-based, prediction-based, and visualization-based)  to validate the effectiveness of our framework on real-world data.

\begin{figure}[htbp]
\vspace{-0.3cm}
    \centering
    \includegraphics[width=0.5\textwidth]{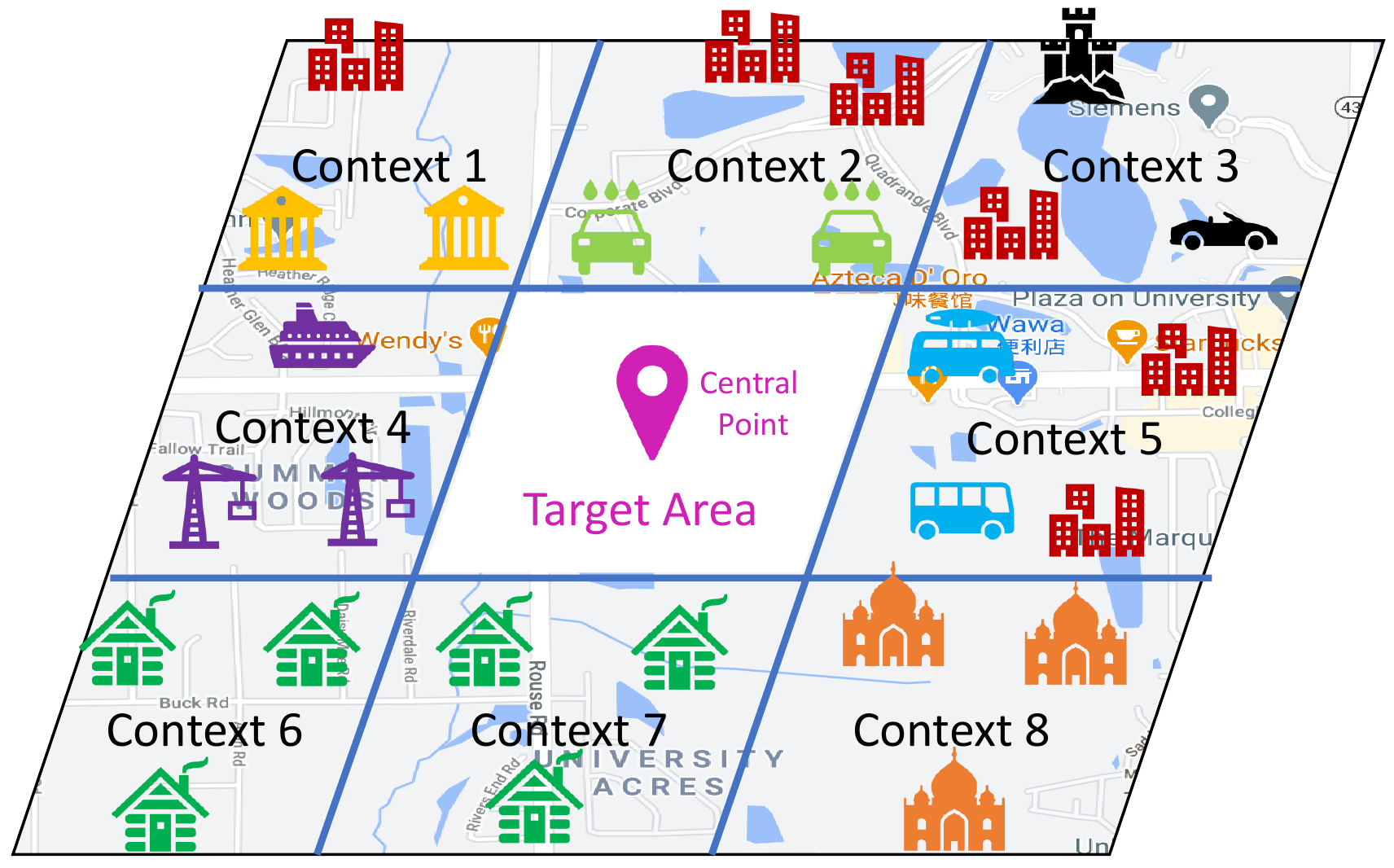}
    \vspace{-0.3cm}
    \caption{The geographical spatial relations between a target area and the surrounding contexts.}
    \vspace{-0.4cm}
    \label{residential_context}
    \vspace{-0.3cm}
\end{figure}

\section{Problem Statement and Framework Overview}

\subsection{Definitions}
\subsubsection{Target Area} refers to an geographical area, where is centered on a geographical location (described by latitude and longitude), and the shape of the area is square. 
% We aims to generate the land-use configuration of the target area.

\subsubsection{Surrounding Contexts} refer to the surrounding squares which wraps the target area from different directions.
The shape of each square in the surrounding contexts is same as the target area.
In our research assumption, we have known the information such as demographic data, social activity, traffic volume, etc of the surrounding contexts.
According to the geographical vicinity and the information of the surrounding contexts, we construct a spatial attributed graph, in which the vertices are the squares of the surrounding contexts and the attributes of each vertex are extracted from the information of each square.
Figure \ref{residential_context} shows the geographical spatial relations between a target area and the surrounding contexts, in which different contexts have different urban utility and characteristics.
Our framework aims to generate the land-use configuration of the target area based on the surrounding contexts.

\subsection{Problem Statement}
As mentioned before, we aim to build up an automated generation framework that generates land-use configuration of the target area based on the surrounding contexts.
Formally, assuming a target area is $R$, the surrounding contexts of $R$ are $[C_1 \sim C_K]$, and the land-use configuration for $R$ is $\mathbf{M}$ that is a longitude-latitude-channel tensor.
Given a spatial attributed graph $G$ that is constructed by extracting explicit features such as traffic condition, economic development, etc from surrounding contexts $[C_1 \sim C_K]$,
we aim to find the mapping function $f: G \rightarrow \mathbf{M}$.
The function takes the spatial attributed graph $G$ as input, and outputs the land-use configuration $\mathbf{M}$.
In this paper, owing to the shape of the target area is square, the number of the squares in the surrounding contexts is determined as $K=8$.

\begin{figure}[!thbp]
    \centering
    \vspace{-0.4cm}
    \includegraphics[width=0.9\linewidth]{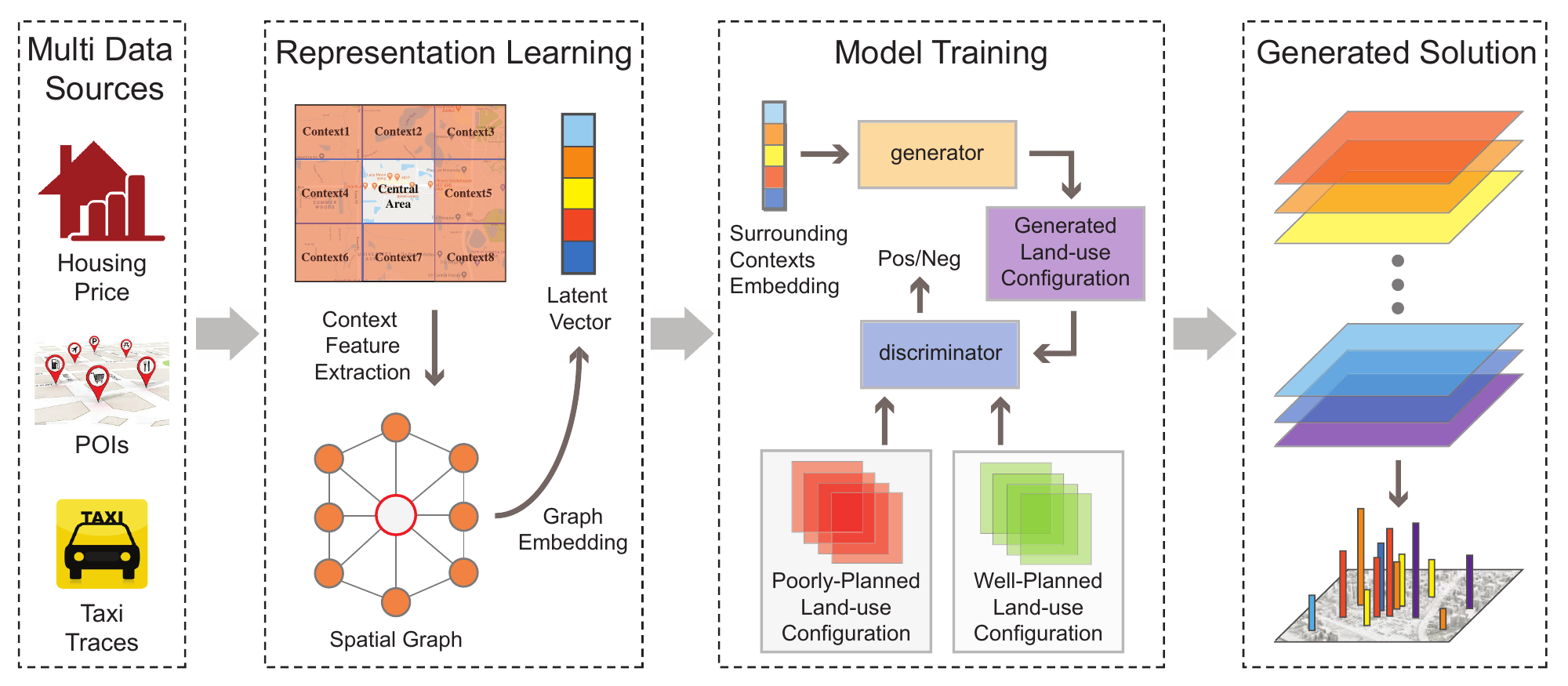}
    \vspace{-0.4cm}
    \caption{An overview of the proposed framework. The proposed framework includes four steps: 1) we first collect multiple data sources such as urban community related data (housing prices), point of interests data, and human mobility data (taxicab GPS traces). We then utilize a spatial graph representation learning module to learn the representations of surrounding contexts. Finally, we develop an adversarial land-use configuration generation model to automate planning and generate recommended configurations. }
    \label{framework}
    \vspace{-0.7cm}
\end{figure}

\subsection{Framework Overview}

Figure \ref{framework} shows an overview of our proposed method (LUCGAN).
This framework has two main phases: 
(i)  surrounding contexts embedding phase;
(ii) land-use configuration generation phase.
In the surrounding contexts embedding phase, we first extract explicit features of the surrounding contexts from multiple aspects, such as value-added space, POI distribution.
Then, we model the eight squares of the surrounding contexts as eight vertices and map the explicit features to the vertices as the corresponding attribute to construct a spatial attributed graph. 
Next, we employ a graph embedding model to preserve the information of the graph into an embedding vector.
Through the above procedures, the final embedding vector represents the whole surrounding contexts.
In the land-use configuration generation phase, we first input the embedding of the contexts into an extended generative adversarial networks (GAN). 
Then, the GAN model learns to formulate the distribution of the  well-planned land-use configurations instead of poorly-planned configurations gradually.
Finally, when the model converges, the extended GAN can produce suitable and desired land-use configurations based on the embeddings of the surrounding contexts.

\section{Automatic planner for land-use configuration}

In this section, we first introduce the strategy to represent surrounding contexts.
Then, we detail how to quantify and evaluate the quality of land-use configurations.
Finally, we develop an automated urban planner based on deep generative adversarial paradigm.

\subsection{Extraction of Explicit Features of  Surrounding Contexts}
The surrounding contexts affect the land-use configuration of a target area.
For instance, if the surrounding contexts own lots of recreational facilities, in order to avoid waste of resources,  we will not plan lots of  recreational buildings in the target area. 
Instead, we prefer to choose other kinds of buildings such as commercial or educational buildings to make the target area coexist with the surrounding contexts in harmony.
Thus, based on the observation, during land-use configuration generation process, it is necessary to take surrounding contexts into consideration. In this paper, we extract the explicit features of surrounding contexts from four aspects:

\begin{enumerate}
    \item \textbf{Value-added Space.}
        Commonly, the variation of housing price reflects the value-added space of one area.
        Thus, we calculate the dynamically changing trend of housing price of the contexts $[C_1 \sim C_8]$ in continuous $t$ months.
         Here, we take the context $C_1$ as an example to explain the calculation process.
         First, we obtain the housing price list among $t$ months.
         Then, we calculate the changing trend of housing price by using the current housing price to subtract the previous housing price.
         So we get the changing trend of $C_1$ as $\mathbf{v}_1=[v_1^1,v^2_1,...,v^{t-1}_1]$, where $v_1^i$ represents the value of the changing trend at $i$-th month.
         Finally, we collect the housing price changing trend of all contexts together. The collected result is denoted as $\mathbf{V} = [\mathbf{v}_1,\mathbf{v}_2,...,\mathbf{v}_8]$, where $\mathbf{V} \in \mathbb{R}^{8 \times {(t-1})}$.
         
    \item \textbf{POI Ratio.}
        Since different POIs provide different services for residents, the ratio of different kinds of POIs can reflect the utility of one area.
        Therefore, we calculate the POI ratio of the contexts $[C_1 \sim C_8]$.
        Here, we take $C_1$ as an example to explain the calculation process.
        First, we count the total number of POI belonging to each POI category in $C_1$ respectively to form a vector.
        Then, we divide each item in the vector by the number of all POIs in $C_1$ to obtain the POI ratio vector, denoted by $\mathbf{r}_1 = [r_1^1,r^2_1,...,r^m_1]$, where $r_1^i$ represents the ratio of i-th POI category and $m$ is the number of POI categories.
        Finally, we collect the POI ratio vector of all contexts together, denoted as $\mathbf{R} = [\mathbf{r}_1,\mathbf{r}_2,...,\mathbf{r}_8]$, where  $\mathbf{R} \in \mathbb{R}^{8\times{m}}$.

    \item \textbf{Public Transportation.}
    Public transportation (i.e. bus, subway) is one of the most important travel modes due to its convenience and economy. 
    We need to consider the public transportation of the contexts $C_1 \sim C_8$.
    Here, we take $C_1$ as an example to show the calculation details.
    To capture the characteristics of public transportation, we extract features based on bus trajectory and bus station data from five perspectives:
      (1) the leaving volume of $C_1$ in one day, denoted by $\mathbf{o}_1^1$;
      (2) the arriving volume of $C_1$ in one day, denoted by $\mathbf{o}_1^2$;
      (3) the transition volume of $C_1$ in one day, denoted by $\mathbf{o}_1^3$;
      (4) the density of bus stop of $C_1$, denoted by $\mathbf{o}_1^4$;
      (5) the average balance of smart card of $C_1$, denoted by $\mathbf{o}_1^5$.
      Thus, the feature vector of $C_1$ can be denoted as $[o_1^1,o^2_1,...,o^5_1]$.
      Finally, we collect the feature vectors of all contexts together.
      The collected result is denoted as $\mathbf{O} = [\mathbf{o}_1,\mathbf{o}_2,...,\mathbf{o}_8]$, where  $\mathbf{O} \in \mathbb{R}^{8 \times 5}$. 
     
     \item \textbf{Private Transportation.}
     Private transportation (i.e. taxi, cab) is another important travel mode for individuals due to its flexibility.
     We extract the features of private transportation of the contexts $[C_1 \sim C_8]$ based on taxi trajectory data from 5 perspectives. 
     Taking $C_1$ as an example, the definitions of the 5 features as follows: 
     (1) the leaving volume of $C_1$ in one day, denoted by $u_1^1$;
     (2) the arriving volume of $C_1$ in one day, denoted by $u_1^2$;
     (3) the transition volume of $C_1$ in one day, denoted by $u_1^3$;
     (4) in $C_1$, the average driving velocity of taxis in one hour, denoted by $u_1^4$;
     (5) in $C_1$, the average commute distance of taxis in one hour, denoted by $u_1^5$;
     Then, the feature vector of private transportation is denoted as $[u_1^1,u^2_1,...,u^5_1]$.
     Finally, we collect the all context features together,  denoted as $\mathbf{U}=[\mathbf{u}_1,\mathbf{u}_2,...,\mathbf{u}_8]$, where  $\mathbf{U} \in \mathbb{R}^{8 \times 5}$.
      
\end{enumerate}

After that, we obtain an explicit feature set from the contexts $C_1 \sim C_8$.
The set contains four kinds of features  $[\mathbf{V},\mathbf{R},\mathbf{O},\mathbf{U}]$, which describes the surrounding contexts from aforementioned perspectives.

\subsection{Constructing Spatial Attributed Graphs with Explicit Features as Node Attributes}

\begin{figure}[htbp]
    \centering
    \vspace{-0.3cm}
    \includegraphics[width=0.23\linewidth]{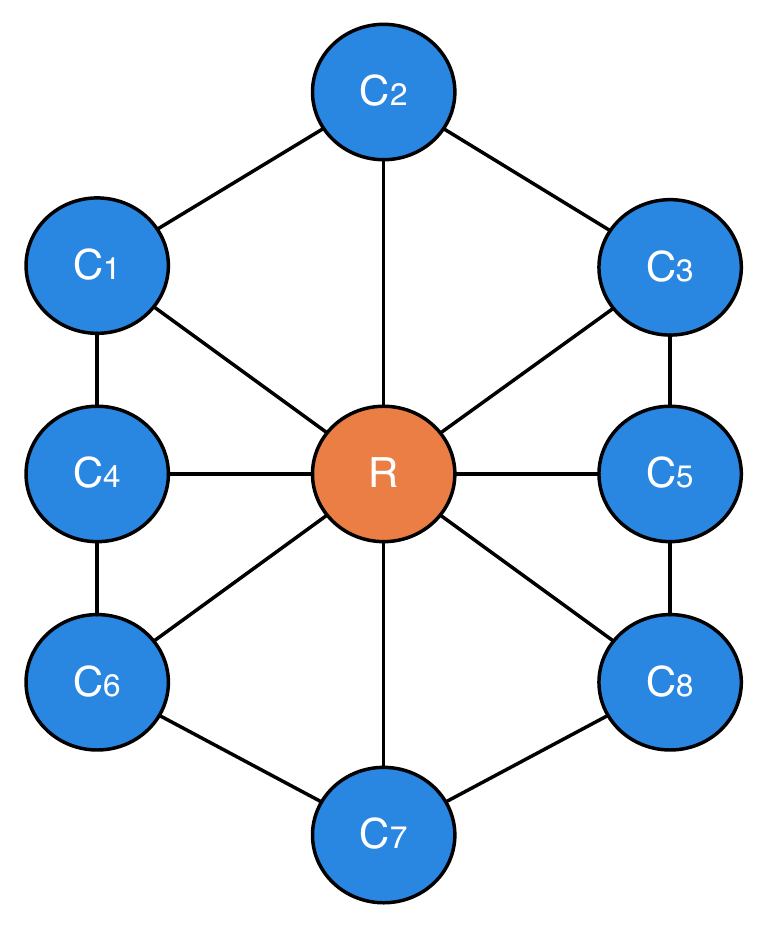}
        \vspace{-0.1cm}
    \caption{The spatial structural relation between a target area and its surrounding contexts. }
    \vspace{-0.4cm}
    \label{spatial_graph}
\end{figure}

The surrounding contexts wrap the target area from different directions, resulting in spatial correlation among areas.
To capture the spatial correlations among the areas, we construct a spatial attributed graph.
Specifically, Figure \ref{spatial_graph} shows the graph structural relation between a target area and its surrounding contexts, where the blue vertices represent the surrounding contexts;
the orange vertex indicates the target area;
the edge between two vertices reflects the spatial connectivity between them.

\begin{figure}[htbp]
    \centering
    \vspace{-0.3cm}
    \includegraphics[width=0.6\linewidth]{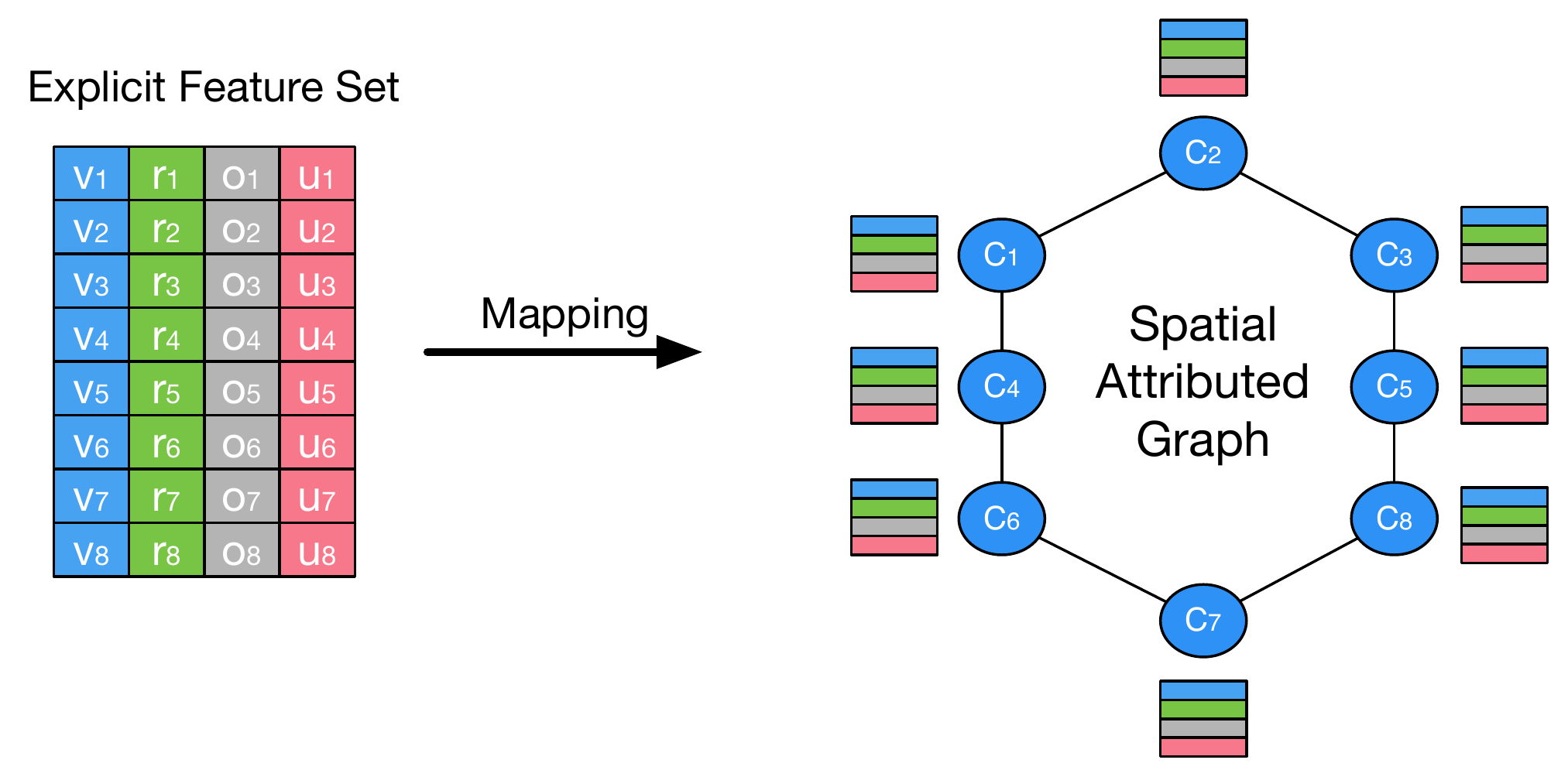}
    \vspace{-0.1cm}
    \caption{The illustration of constructing a spatial attributed graph: Each feature vector is mapped to the corresponding vertex by a column-wise strategy.}
    \vspace{-0.5cm}
    \label{graph_feature}
\end{figure}

Then, we map the explicit features $[\mathbf{V},\mathbf{R},\mathbf{O},\mathbf{U}]$ to the spatial graph structure as the corresponding node attributes.
Figure \ref{graph_feature} expresses the mapping process.
The final spatial attributed graph not only reflects the spatial correlation among different context squares but also depicts the utility characteristics of each square.

\subsection{Learning Representation of  Spatial Attributed Graphs}

\begin{figure}[t]
    \centering
    \vspace{-0.3cm}
    \includegraphics[width=0.6\linewidth]{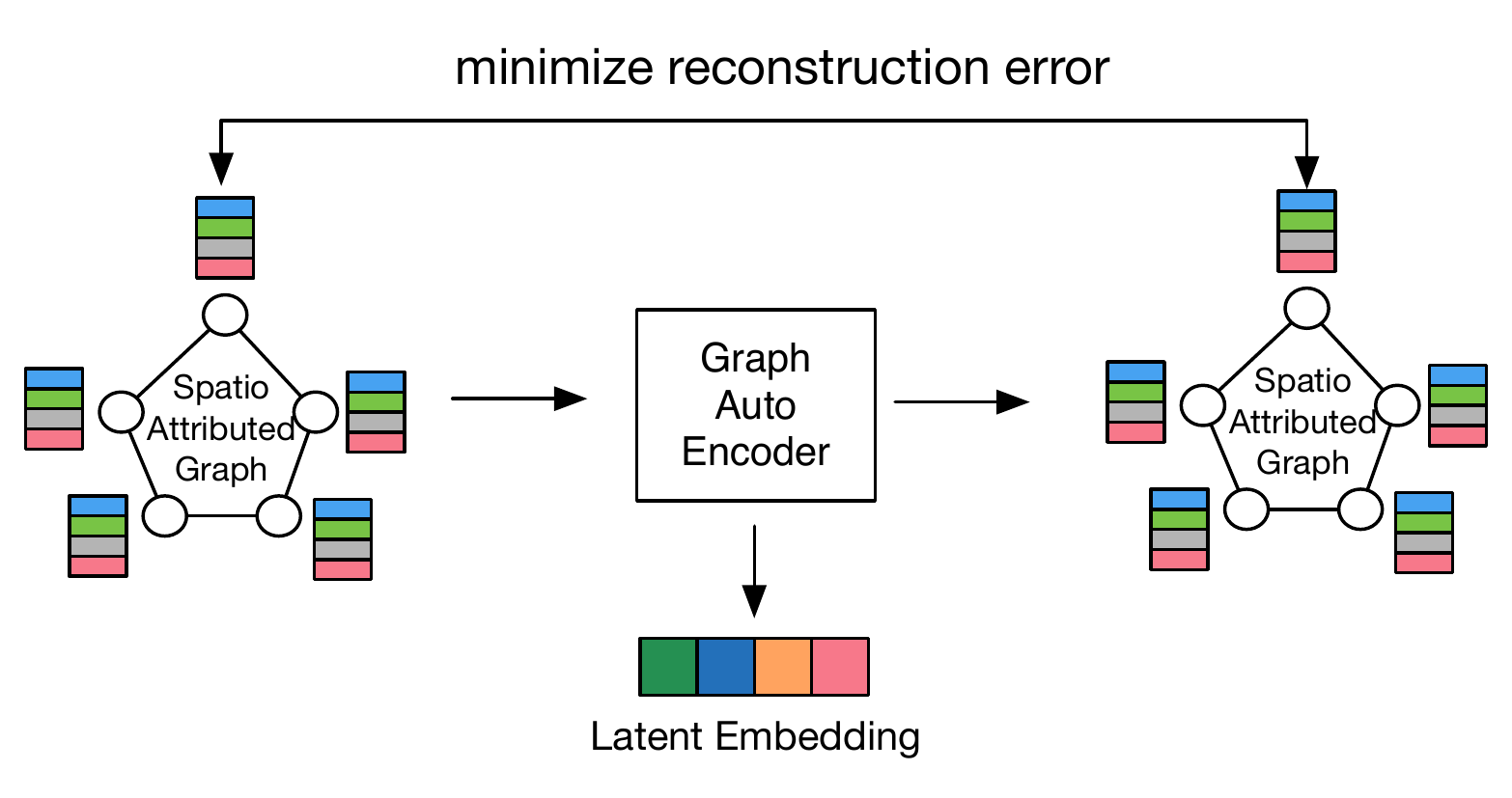}
    \vspace{-0.6cm}
    \caption{The proposed representation learning module to obtain surrounding  context representations by minimizing the reconstruction loss of spatial attributed graphs.}
    \vspace{-0.5cm}
    \label{gae_feature}
\end{figure}

Figure \ref{gae_feature} shows the spatial representation learning framework that preserves explicit features and spatial relations of the spatial attributed graph into a low-dimensional vector. 
Formally, we denote the spatial attributed graph as $G=(\mathbf{X},\mathbf{A})$,  where $\mathbf{A} \in \mathbb{R}^{8\times 8}$ is the adjacency matrix that expresses the accessibility among different nodes;
$X\in \mathbb{R}^{8\times(t+m+9)}$ is the feature matrix of the graph, here, $\mathbf{X}=[\mathbf{V},\mathbf{R},\mathbf{O},\mathbf{U}]$. 
In order to get the latent graph embedding $\mathbf{z}$,
we minimize the reconstruction loss between the original graph $G$ and the reconstructed graph $\widehat{G}$ through the encoding-decoding paradigm.

The encoding part has two Graph Convolutional Network (GCN) layers.
The first GCN layer takes $\mathbf{X}$ and $\mathbf{A}$ as input and outputs the feature matrix of low-dimensional space $\widehat{\mathbf{X}}$.
Thus, the encoding module can be formulated as:
\begin{equation}
    \widehat{\mathbf{X}} = GCN_1(\mathbf{X},\mathbf{A})=RELU(\widehat{\mathbf{D}}^{-\frac{1}{2}}\mathbf{A}\widehat{\mathbf{D}}^{-\frac{1}{2}}\mathbf{XW}_{1})
\end{equation}
where $\widehat{\mathbf{D}} \in \mathbb{R}^{8\times 8}$ is the diagonal degree matrix, $\mathbf{W}_1 \in \mathbb{R}^{8 \times M}$ is the weight matrix of the $GCN_1$ where $M$ is the output dimension of the layer, and the whole layer is activated by $RELU$ function.
The second GCN layer takes $\widehat{\mathbf{X}}$ and $\mathbf{A}$ as input and then outputs the mean value $\bm{\mu}$ and the variance value $\bm{\delta}^2$ of normal distribution.
So the calculation process of the second GCN layer can be formulated as:
\begin{equation}
    \bm{\mu},log(\bm{\delta}^2) = GCN_2(\mathbf{\widehat{X}},\mathbf{A}) = \widehat{\mathbf{D}}^{-\frac{1}{2}}\mathbf{A}
    \widehat{\mathbf{D}}^{-\frac{1}{2}}\widehat{\mathbf{X}}\mathbf{W}_2
\end{equation}
where $\mathbf{W}_2 \in \mathbb{R}^{M \times H}$ is the weight matrix of $GCN_2$.
Here, $H$ is the output dimension of the $GCN_2$ layer.
Next, we use the reparameterization trick to obtain the latent representation $\mathbf{z} \in \mathbb{R}^{8\times H}$:

\begin{equation}
    \mathbf{z}=\bm{\mu}+\bm{\delta} \times \epsilon
\end{equation}
where $\epsilon \sim \mathcal{N}(0,1)$.

The decoding module takes the $\mathbf{z}$ as input and then outputs the reconstructed adjacent matrix $\widehat{\mathbf{A}}$.
Hence, the decoding step can be formulated as:
\begin{equation}
    \widehat{\mathbf{A}} = \sigma(\mathbf{z}\mathbf{z}^T)
\end{equation}
where $\sigma$ represents the decoding layer activated by sigmoid function. 
Moreover, $\mathbf{z}\mathbf{z}^T$ can be converted to $\left \|\mathbf{z}\right \| \left \|\mathbf{z}^T\right\| \cos\theta$.
The inner product operation is beneficial to capture the spatial correlation among different contexts.

During the training phase, we minimize the joint loss function $\mathcal{L}$, denoted as:
\begin{equation}
    % \begin{split}
        \mathcal{L} = \sum \limits_{i=1}^{N} \underbrace{
        KL[q(\mathbf{z}|\mathbf{X},\mathbf{A}) || p(\mathbf{z})]
        }_{\text{KL Divergence between $q(.)$ and $p(.)$}}
        +
        \overbrace{
        \sum_{j=1}^{S} \left \| \mathbf{A}-\widehat{\mathbf{A}} \right \|^2 
        }^{\text{Loss between $\mathbf{A}$ and $\widehat{\mathbf{A}}$}}
    \label{equ:loss}
\end{equation}
where $N$ is the dimension of $\mathbf{z}$;
$S$ is the total number of the vertices in $\mathbf{A}$;
$q$ represents the real distribution of $\mathbf{z}$;
$p$ represents the prior distribution of $\mathbf{z}$.
$\mathcal{L}$ includes two parts, the first part is the Kullback-Leibler divergence between the standard prior distribution $\mathcal{N}(0,1)$ and the distribution of $\mathbf{z}$, and the second part is the squared error between $\mathbf{A}$ and $\widehat{\mathbf{A}}$.
The training process try to make $\widehat{\mathbf{A}}$ get close to $\mathbf{A}$ and let the distribution of $\mathbf{z}$ get close to $\mathcal{N}(0,1)$.
When the model converges, $\mathbf{z}$ contains all information of the surrounding contexts.

\subsection{Land-use Configuration Quantification and Quality Measurement}

\begin{figure}[!thbp]
    \centering
    \vspace{-0.3cm}
    \includegraphics[width=0.85\linewidth]{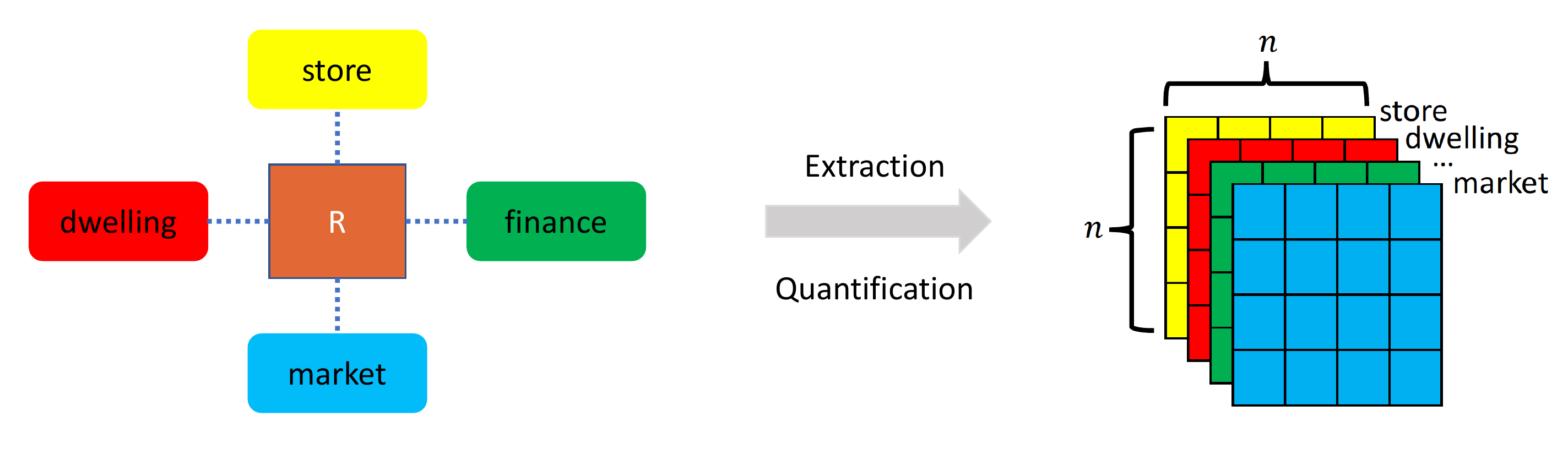}
    \vspace{-0.2cm}
    \caption{ The construction of longitude-latitude-channel configuration tensor. Specifically, we first collect the information such as dwelling, market, etc of an area. Then, we extract and quantify the information of the area as a multi-channel tensor, where the value of each entry is the number of POIs with respect to a specific category in a specific latitude range and longitude range.}
    \label{poi_dis}
    \vspace{-0.3cm}
\end{figure}

Land-use configuration indicates the location of different kinds of POIs in one area.
To make a machine perceive and understand the configuration, we construct a longitude-latitude-channel tensor as the format of the configuration, where one channel denotes one POI category and the whole tensor represents the POI distribution in the area. 
Figure \ref{poi_dis} shows the construction process of the longitude-latitude-channel configuration tensor. 
We first divide an unplanned target area into $n\times n$ squares.
Then we count the number of POIs belonging to each POI category in each square entry and fill the number into the corresponding entry respectively.
In this way, we obtain the land-use configuration tensor. 
If we pick up one channel from the tensor, we can learn about the POI distribution of the corresponding POI category in the whole area.

Owing to we expect the generation framework can generate well-planned land-use configuration, the next big question is how to evaluate the quality of the configuration? 
In the classical urban planning domain, there are no general evaluation standards since the complexity of urban systems.
To make our framework can produce the land-use configurations that people satisfied with, we provide a quality hyper-parameter $Q$ to evaluate the quality of land-use configurations.  
In our experiment, $Q$ is the combination of the POI diversity and the check-in frequency.
Formally, we first count the total frequency number of mobile check-in events of an area, which reflect the social activity intensity, denoted by $freq$. 
Then, we calculate the total number of different POI categories of the area as the POI diversity, which depicts the completeness of urban functions, denoted by $div$.
Next, we incorporate the two indicators together by the formula $Q=\frac{2\times freq \times div}{freq+div}$ \cite{wang2018ensemble}.
If $Q > \text{threshold}$, the configuration of the area is regarded as a well-planned configuration, 
otherwise, it is justified as a poorly-planned configuration.
Here, the value of $\textit{threshold}$ is determined by given requirements.

\subsection{Land-use Configuration Generative Adversarial Networks}
Recently, Generative Adversarial Networks (GANs) achieve tremendous achievements and reveal strong imaginative and generative abilities.
It motivates us to formulate the land-use configuration generation task into the learning paradigm of GAN.

\begin{figure*}[t]
% \vspace{-0.3cm}
    \centering
    \includegraphics[width=0.85\linewidth]{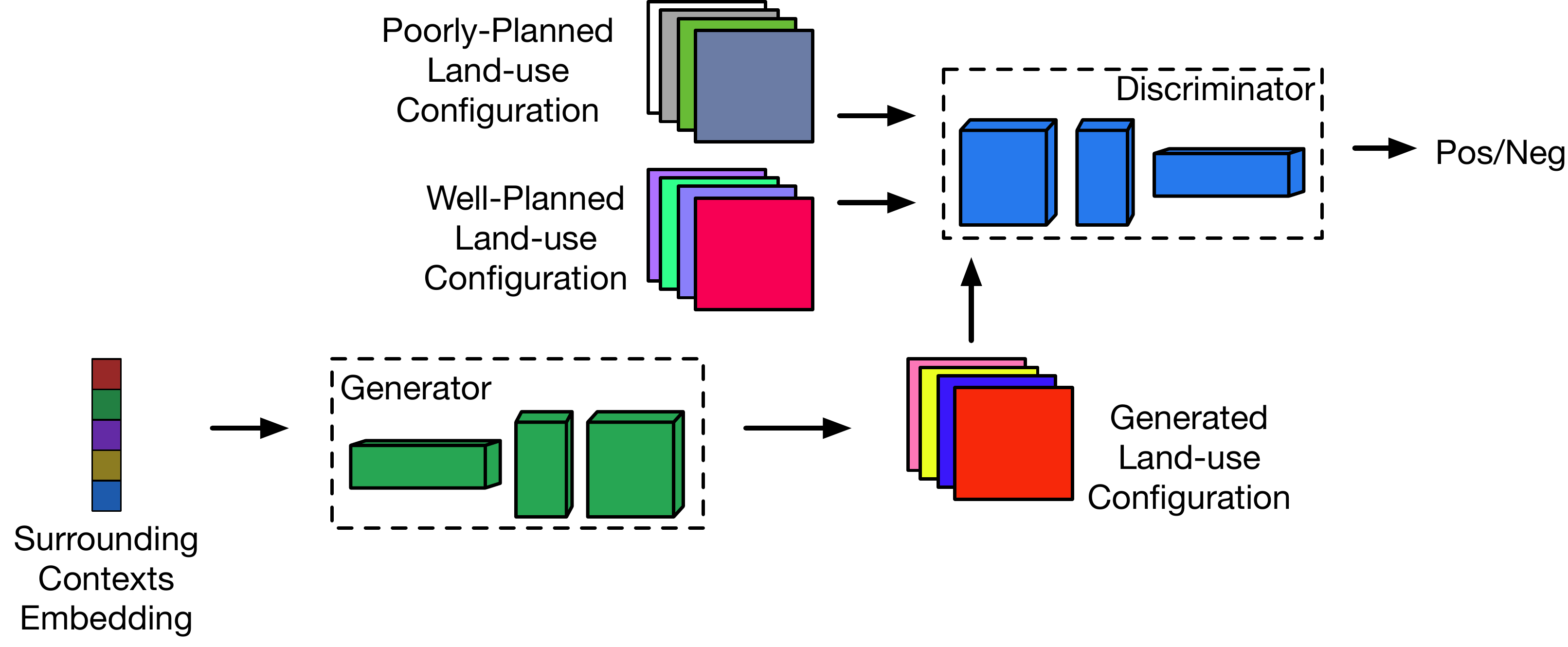}
    \vspace{-0.3cm}
    \caption{The model structure of LUCGAN}
    \vspace{-0.3cm}
    \label{LUCGAN}
\end{figure*}

In our preliminary version~\cite{wang2020reimagining}, we propose a land-use configuration GAN (LUCGAN), and the network structure of LUCGAN as illustrated in Figure ~\ref{LUCGAN}.
In LUCGAN, the generator generates land-use configuration based on the embeddings of surrounding contexts. 
The discriminator provides feedback to the generator for generating configurations close to well-planned configurations instead of poorly-planned configurations.

The Algorithm \ref{alg:lucgan} shows the training process of LUCGAN.
Specifically, in one training iteration, we first update the parameters of the discriminator for $\kappa$ times, then learn the parameters of the generator for 1 time based on the current discriminator. 
For the updating process of the discriminator, we sample $m$ well-planned configurations, surrounding context embeddings, and poorly-planned configurations respectively. 
We utilize them to maximize the loss function illustrated in line 10 of Algorithm \ref{alg:lucgan}.
Intuitively, we expect the discriminator to provide positive feedback for well-planned configurations, and negative feedback for poorly-planned and generated configurations.
In this way, the discriminator improves the distinguishing ability for land-use configurations.
For the updating process of the generator, we sample $m$ surrounding context embeddings firstly.
Then we minimize the loss function shown in line 14 of Algorithm \ref{alg:lucgan}.
Intuitively, we aim to utilize the discriminator to improve the generative ability of the generator for producing data structures similar to well-planned configurations.

\begin{figure*}[t]
% \vspace{-0.3cm}
    \centering
    \includegraphics[width=0.9\linewidth]{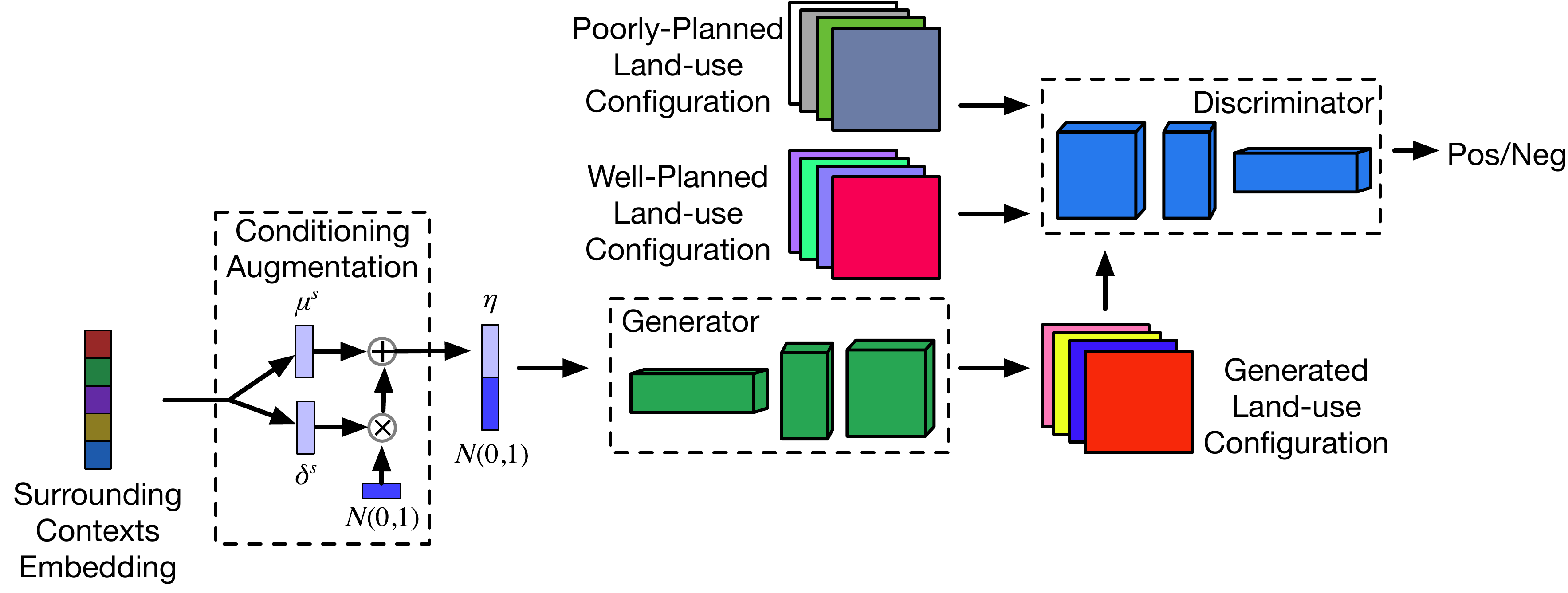}
    \vspace{-0.3cm}
    \caption{The model structure of LUCGAN$^+$}
    \vspace{-0.6cm}
    \label{LUCGAN+}
\end{figure*}

\begin{algorithm}[hbtp] 
  \tcp{start training.}
  \For{number of training iterations}{
  \tcp{update discriminator firstly.}
  \For{$\kappa$ steps}{
    Sample $m$ well-planned land-use configuration samples $\left \{ \mathbf{E}^1,\mathbf{E}^2,...,\mathbf{E}^m \right \}$.\\
    Sample $m$ surrounding context embeddings $\left \{ \mathbf{z}^1,\mathbf{z}^2,...,\mathbf{z}^m \right \}$. \\
    Generate land-use configurations by generator,  $\left \{ \mathbf{F}^1,\mathbf{F}^2,...,\mathbf{F}^m \right \}$, $\mathbf{F}^i=G(\mathbf{z}^i)$.\\
    Sample $m$ poorly-planned land-use configurations$\left \{ \mathbf{T}^1,\mathbf{T}^2,...,\mathbf{T}^m \right \}$. \\
    Update the discriminator by maximizing the following loss:
    ~\\
    $\bigtriangledown_{\theta_d} \frac{1}{m}\sum_{i=1}^{m} [ log(D(\mathbf{E}^i)) + log(D(1-\mathbf{F}^i))+ log(D(1-\mathbf{T}^i)) ].$
    ~\\
  }
  \tcp{update generator secondly.}
    Sample $m$ context information embedding samples
    $\left \{ \mathbf{z}^1,\mathbf{z}^2,...,\mathbf{z}^m \right \}$. \\
    Update the generator by minimizing the following loss:\\
    $\bigtriangledown_{\theta_g} \frac{1}{m} \sum_{i=1}^{m} log(1-D(G(\mathbf{z}^i))).$
    ~\\
  }
 
  \caption{The training process of LUCGAN. Here,$G$ denotes the generator, $D$ denotes the discriminator. We adjust the hyperparameter $\kappa$ to change the updating frequencies of the parameters of the discriminator.}
  \label{alg:lucgan}
\end{algorithm}

\begin{algorithm}[hbtp] 
  \tcp{start training.}
  \For{number of training iterations}{
  \tcp{Conditioning Augmentation.$\bm{\mu}^s, \bm{\delta}^s$ are mean and variance respectively. $\mathbf{W}_{()},\mathbf{b}_{()}$ are weight and bias of the corresponding item respectively. $\epsilon^{()}$ indicates that sampling a vector from normal distribution $\mathcal{N}(0,1)$.} 
  Sample $m$ context information embedding samples $\mathbf{Z} = \left \{ \mathbf{z}^1,\mathbf{z}^2,...,\mathbf{z}^m \right \}$. \\
  $\bm{\mu}^s = RELU(\mathbf{W}_{\mu}\cdot\mathbf{Z}+\mathbf{b}_{\mu})$;\\
  $\bm{\delta}^s = RELU(\mathbf{W}_{\delta}\cdot\mathbf{Z}+\mathbf{b}_{\delta})$.\\
  $\bm{\eta} = Concatenate((\bm{\mu}^s+\bm{\delta}^s\cdot \epsilon^s),\epsilon^c)$.\\
  \tcp{update discriminator firstly}
  \For{$\kappa$ steps}{
    Sample $m$ well-planned land-use configurations $\left \{ \mathbf{E}^1,\mathbf{E}^2,...,\mathbf{E}^m \right \}$.\\
    Collect $m$ vectors as context embeddings thorough  line 6 $\left \{ \bm{\eta}^1,\bm{\eta}^2,...,\bm{\eta}^m \right \}$. \\
    Generate land-use configurations by generator,  $\left \{ \mathbf{F}^1,\mathbf{F}^2,...,\mathbf{F}^m \right \}$, $\mathbf{F}^i=G(\bm{\eta}^i)$.\\
    Sample $m$ poorly-planned land-use configurations$\left \{ \mathbf{T}^1,\mathbf{T}^2,...,\mathbf{T}^m \right \}$. \\
    Update the discriminator by maximizing the following loss:
    ~\\
    $\bigtriangledown_{\theta_d} \frac{1}{m}\sum_{i=1}^{m} [ log(D(\mathbf{E}^i)) + log(D(1-\mathbf{F}^i))+ log(D(1-\mathbf{T}^i)) ].$
    ~\\
  }
  \tcp{update generator secondly.$KL$ means Kullback-Leibler divergence}
    Collect $m$ vectors as contexts embeddings thorough  line 6 $\left \{ \bm{\eta}^1,\bm{\eta}^2,...,\bm{\eta}^m \right \}$. \\
    Update the generator by minimizing the following loss:\\
    $\bigtriangledown_{\theta_g} \frac{1}{m} \sum_{i=1}^{m} log(1-D(G(\bm{\eta}^i))) + KL(\mathcal{N}(\bm{\mu}^s,\bm{\delta}^s) \left |  \right | \mathcal{N}(0,1)).$
    ~\\
  }
 
  \caption{The training process of LUCGAN$^+$. Here,$G$ denotes the generator, $D$ denotes the discriminator. We adjust the hyperparameter $\kappa$ to change the updating frequencies of the parameters of the discriminator.}
  \label{alg:lucgan+}
\end{algorithm}

However, the embeddings of the surrounding contexts come from a feature space constructed by spatial attributed graphs.
Owing to the small number of graphs, the distribution of them in the feature space is sparse and discrete, which causes the learning process of the GAN model unstable. 
To overcome this limitation and improve model performance, we propose an enhanced framework, namely LUCGAN$^+$, and the network structure as shown in Figure \ref{LUCGAN+}.
Compared with Figure \ref{LUCGAN}, we add a conditioning augmentation module ~\cite{zhang2017stackgan} into our framework.
Specifically, we first assume the prior distribution of the surrounding contexts embeddings is a normal distribution.
Then, we estimate the mean and variance of the distribution based on the original embeddings.
Next, we sample a vector from the distribution and combine it with a vector sampled from standard normal distribution as the input vector of the model.
This process improves the model performance because it mitigates the discreteness and sparsity of original graphs in the feature space.

In addition, owing to the differences of the model structure between LUCGAN and LUCGAN$^+$, we customize a new training algorithm for LUCGAN$^+$ as shown in Algorithm \ref{alg:lucgan+}. 
Compared with Algorithm \ref{alg:lucgan}, there are two improvements: (1) conditioning augmentation module (line 4 ~ line 7 in Algorithm \ref{alg:lucgan+}); (2) loss function of the generator (line 19 in Algorithm \ref{alg:lucgan+}).
For the conditioning augmentation module, we calculate the mean $\bm{\mu}^s$ and the variance $\bm{\delta}^s$ based on original surrounding context embeddings respectively.
Then, we utilize reparametrization trick to sample a vector from normal distribution $\mathcal{N}(\bm{\mu}^s, \bm{\delta}^s)$, and concatenate the vector with a vector sampled from normal distribution $\mathcal{N}(0,1)$ as the surrogate context embeddings $\bm{\eta}$.
For the learning process of the discriminator, the main logic is the same as Algorithm \ref{alg:lucgan}, we only replace the surrounding context embeddings $\mathbf{z}$ with $\bm{\eta}$.
For the loss of the generator, besides improving the generative capability of the generator, we also minimize the Kullback-Leibler (KL) divergence between $\mathcal{N}(\bm{\mu}^s, \bm{\delta}^s)$ and $\mathcal{N}(0,1)$, which enhances the smoothness of the surrounding context embeddings in the feature space and avoids overfitting.

\section{Experiment Results}

In this section, we conduct extensive experiments and case studies to answer the following questions:
\textbf{Q1.} Is our proposed automatic planner effective for generating land-use configurations?
\textbf{Q2.} We split an area into $n\times n$ squares for quantifying land-use configurations. What is the influence of the square size for generating configurations?
\textbf{Q3.} What are the differences between the contexts of well-planned configurations poorly-planned configurations?
\textbf{Q4.} What are the differences of land-use configurations generated by our framework when facing with different planning goals?
\textbf{Q5.} What does the generated result for each POI category look like in a generated land-use configuration?

\begin{table}[htbp]
\vspace{-0.3cm}
\small
\centering
\setlength{\abovecaptionskip}{0.cm}
\caption{POI category list}
\setlength{\tabcolsep}{1mm}{
\begin{tabular}{cccccc}  
\toprule
 code  & POI category & code & POI category  \\  
\midrule       
  0  & road & 10 & tourist attraction \\
  1 & car service & 11 & real estate \\
  2 & car repair & 12 & government place \\
  3 & motorbike service & 13 & education \\
  4 & food service & 14 & transportation \\
  5 & shopping & 15 & finance\\
  6 & daily life service & 16 & company\\
  7 & recreation service & 17 & road furniture\\
  8 & medical service & 18 & specific address \\
  9 & lodging  & 19 & public service\\
\bottomrule
\end{tabular}}
\label{poi_lists}
\vspace{-0.5cm}
\end{table}

\subsection{Data Description}

We use the following datasets for evaluation:
 \textbf{Residential Community:} 
    The residential community dataset contains 2990 residential communities in Beijing \footnote{http://www.soufun.com/}.
    Each community is centered by a geographic point (described by latitude and longitude).
\textbf{POI:} The POI dataset includes 328668 POIs in Beijing ~\footnote{https://www.openstreetmap.org/}. Each POI item includes latitude, longitude, and the corresponding POI category.
    Table \ref{poi_lists} shows the detailed information of POI category.
\textbf{Taxi Trajectories:}
    The taxi trajectories are collected from a Beijing taxi company ~\footnote{https://www.microsoft.com/en-us/research/publication/t-drive-trajectory-data-sample/}. Each trajectory contains trip ID, distance (m), travel time (s), average speed (km/h), pick-up and drop-off time, pick-up and drop-off point.
\textbf{Public Transportation:}
    The public transportation dataset includes bus transactions in Beijing from 2012 to 2013, which contains 718 bus lines, 1734247 bus trips ~\footnote{https://www.beijingcitylab.com/data-released-1/data1-20/}.
\textbf{Housing Price:}
    The housing price dataset is collected from a Chinese real estate website \footnote{http://www.soufun.com/}, which contains the housing price of residential communities of Beijing from 2011 to 2012. 
\textbf{Check-In:}
    The check-in dataset contains the Weibo \footnote{https://open.weibo.com/wiki/2/place/pois/add\_checkin} check-in records in Beijing from 2011 to 2013.
    The data format of one record is: longitude, latitude, check-in time and check-in place.

\subsection{Evaluation Metrics}
We aim to generate land-use configurations that are similar to well-planned configurations.
To evaluate the generative performance, we calculate the difference between the distribution of well-planned configurations $Y$ and the distribution of generated configurations $\widehat{Y}$.
The less distribution difference is, the better generative performance will be.
\begin{enumerate}
    \item \textbf{Kullback-Leibler (KL) Divergence:} 
    $\text{KL}(Y||\widehat{Y}) = \sum Y(x) \cdot \text{ln} \frac{Y(x)}{\widehat{Y}(x)}$, where $x$ is a test sample. 
    \item \textbf{Jensen-Shannon (JS) Divergence:}
 $\text{JS}(Y||\widehat{Y}) = \frac{1}{2}\text{KL}(Y||\frac{Y+\widehat{Y}}{2})+\frac{1}{2}\text{KL}(\widehat{Y}||\frac{Y+\widehat{Y}}{2})$.
%  Compared with KL divergence, JS divergence owns symmetry.
    \item \textbf{Hellinger Distance (HD) :} $HD(Y||\widehat{Y}) =
    \frac{1}{\sqrt{2}}\left \| \sqrt{Y}-\sqrt{\widehat{Y}} \right \|_2.
    $
    \item \textbf{Wasserstein Distance (WD) :} $WD(Y||\widehat{Y}) = inf_{\gamma \sim \Gamma(Y,\widehat{Y})} \mathbb{E}_{(x,y) \sim \gamma}\left \| x-y \right \|$, where $\Gamma(Y,\widehat{Y})$ is a set of joint distribution between $Y$ and $\widehat{Y}$; $\gamma$ is a joint distribution of $\Gamma$; $x, y$ are two samples sampled from $\gamma$; $\mathbb{E}||(.)||$ is the expectation of distances between any two samples.
\end{enumerate}

% Because evaluating the quality of the urban land-use configuration is an open question, there is no standard measurement. 
% In this paper, we evaluate the quality of generated planning solution from multiple aspects to express the effectiveness of our framework:
% (1)\textbf{Scoring Model}.
% We build a random forest model based on the excellent and terrible land-use configuration plans.
% The model is capable of giving higher scores for excellent land-use configuration plans and provide lower scores for terrible plans.
% When we get the generate land-use configuration solutions, the scoring model can be utilized to quantify the quality of the generated solutions.
% (2) \textbf{Visualization}. 
% In order to explore the generated solutions, we select one representative sample to visualize from multiple aspects.
% We can observe the solutions directly in this way.
% It is helpful to learn the difference between our planner and other baselines .

\subsection{Baseline Methods}
We compare the performance of our journal version framework (LUCGAN$^+$) against the following baseline models:
\begin{enumerate}
    \item \textbf{DCGAN}: is an extension for traditional GAN, which utilizes  convolutional layer and convolutional transpose layer in the generator and discriminator respectively ~\cite{radford2015unsupervised}. 
    
    \item \textbf{WGAN}: is a new GAN training framework, which improves the stability of learning and provides meaningful learning curve for debugging and hyperparameter adjustment ~\cite{pmlr-v70-arjovsky17a}.  

    \item \textbf{WGAN$^{\bm{GP}}$}: utilizes gradient penalty to replace clipping weights of WGAN, which enhances the performance of WGAN further ~\cite{gulrajani2017improved}.
    
    \item \textbf{LUCGAN}: is the conference version of our land-use configuration GAN, which is capable of generating the configurations based on the surrounding contexts ~\cite{wang2020reimagining}.
\end{enumerate}

To further study the generated land-use configurations, we adopt two new methods: \textbf{scoring model} and \textbf{visualization}.
For the scoring model, we train a machine learning model to learn the scoring criteria that provides high score for well-planned configurations and low score for poorly-planned configurations. After we obtain all testing samples, the model can be used to evaluate the quality of generated results. 
For the visualization, we visualize the generated results in heat map, pie chart, 3d-bar chart for checking the POI distribution.
We conduct all experiments on a x64 machine with Intel i9-9920X 3.50GHz CPU, 128GB RAM and Ubuntu 18.04.

\subsection{Hyperparameters and Reproducibility}
In our experiments,
first, to obtain the embedding of surrounding contexts (section 3.3), we employ a VGAE ~\cite{kipf2016variational} composed of an encoder and a decoder.
The encoder contains three graph convolutional neural layers.
The decoder only has one reconstructed layer.
We perform Adaptive Moment Estimation (Adam) to optimize the VGAE model with a learning rate of 0.005 for 300 epochs.
The dimension of surrounding contexts' embedding is set to 100.
Second, to quantify the quality of land-use configurations (section 3.4), we set the value of the hyper-parameter $Q$ to 0.5.
Third, our planner LUCGAN$^{+}$ consists of a generator and a discriminator (section 3.5). 
We optimize the generator by Adam with a learning rate of 0.0001.
We perform Stochastic Gradient Descent (SGD) to optimize the discriminator with a learning rate of 0.0001 and a momentum of 0.95.
The whole optimizing process continues for 50 epochs.
To make other researchers easily reproduce our experiments, we release the code and data by Dropbox ~\footnote{https://www.dropbox.com/sh/16pk55efb9fzm2j/AACsosXxHtfQKXKjmL0NrOn1a?dl=0}.

\subsection{Overall Performance (Q1)}
\begin{figure*}[!thb]
\vspace{-0.4cm}
\setlength{\abovecaptionskip}{-2pt} 
	\centering
	\subfigure[KL Divergence]{\label{fig:overall_kl}\includegraphics[width=0.4\linewidth]{{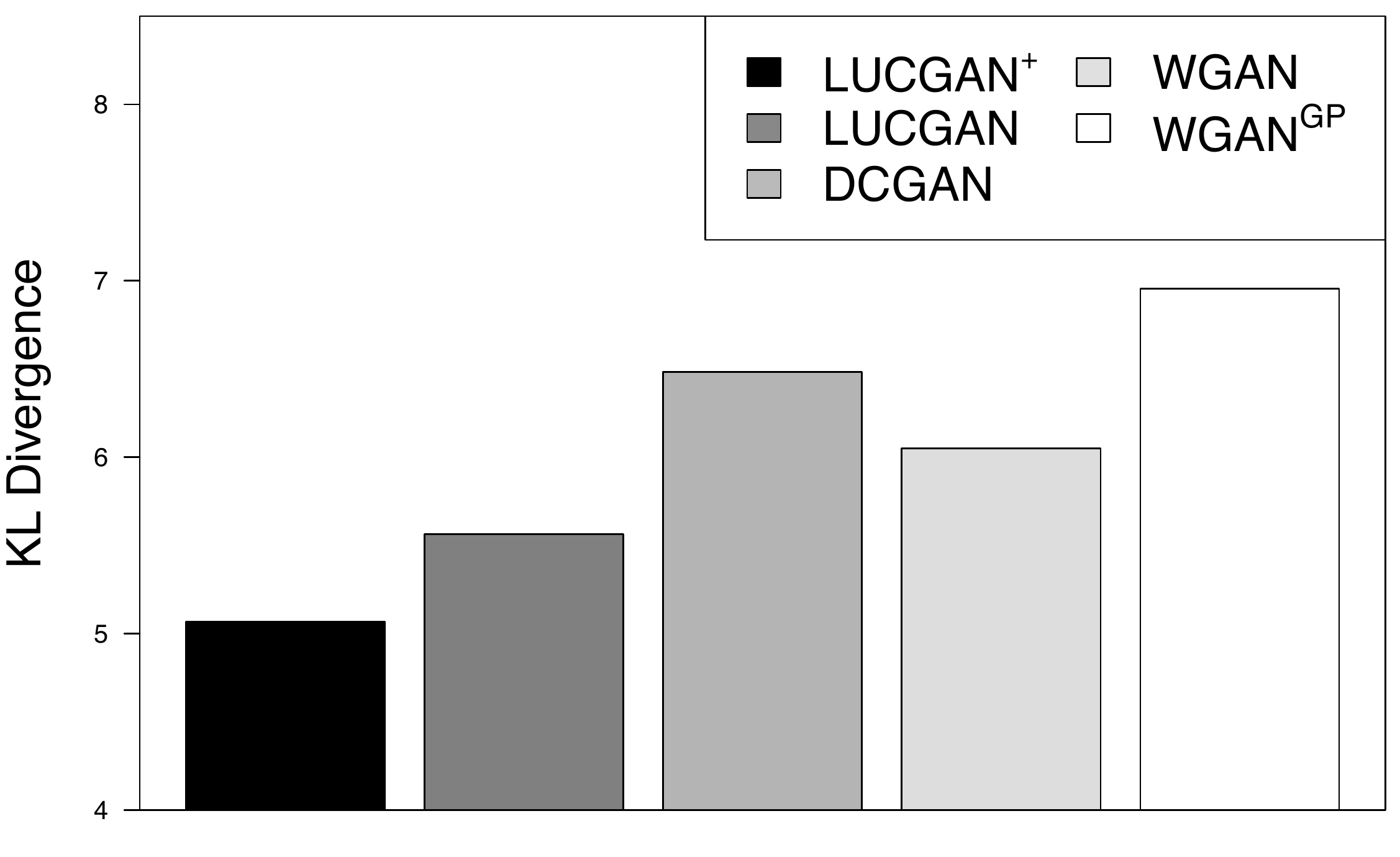}}}
	\subfigure[JS Divergence]{\label{fig:overall_js}\includegraphics[width=0.4\linewidth]{{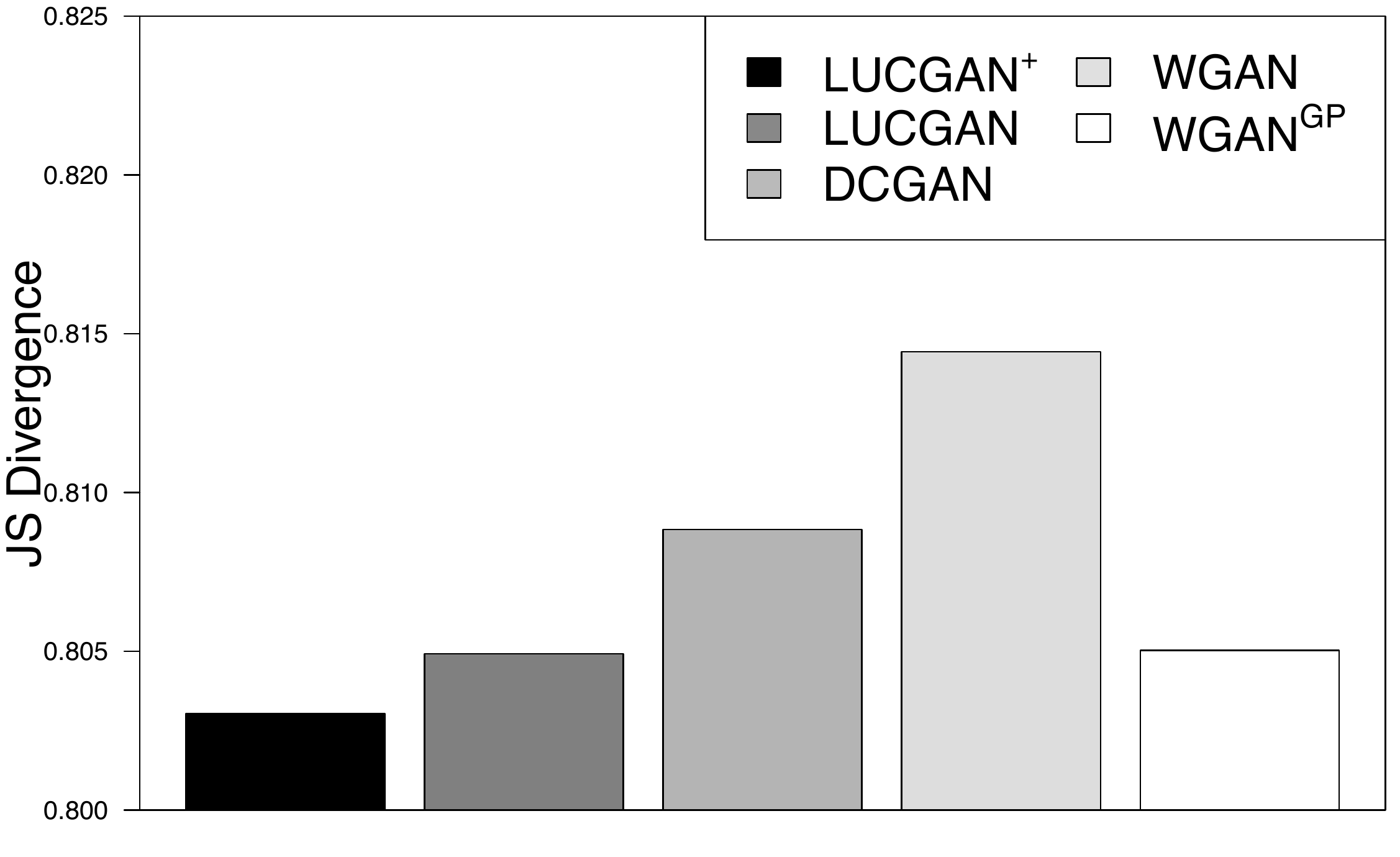}}}

	\subfigure[Hellinger Distance]{\label{fig:overall_hd}\includegraphics[width=0.4\linewidth]{{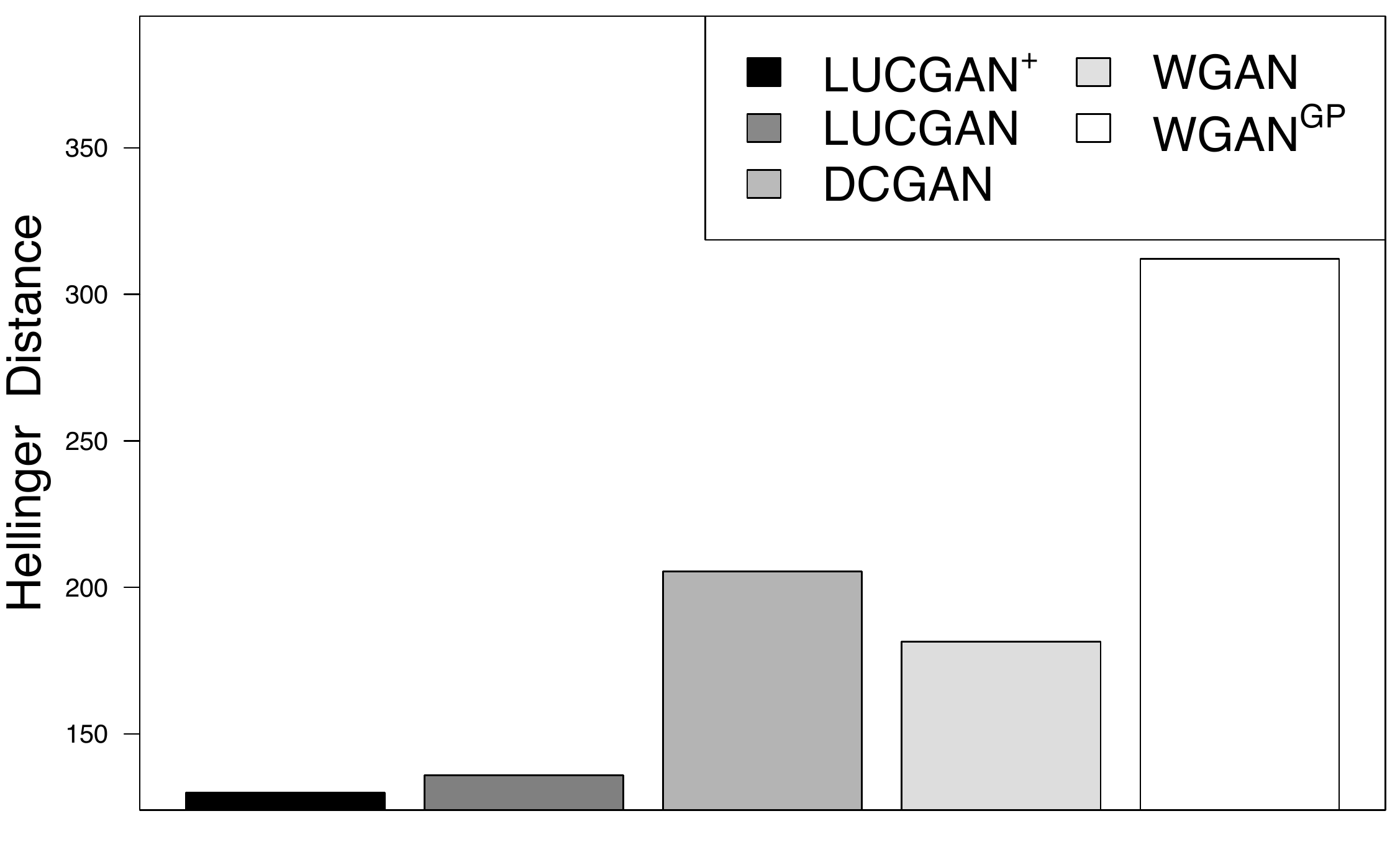}}}
	\subfigure[Wasserstein Distance]{\label{fig:overall_wd}\includegraphics[width=0.4\linewidth]{{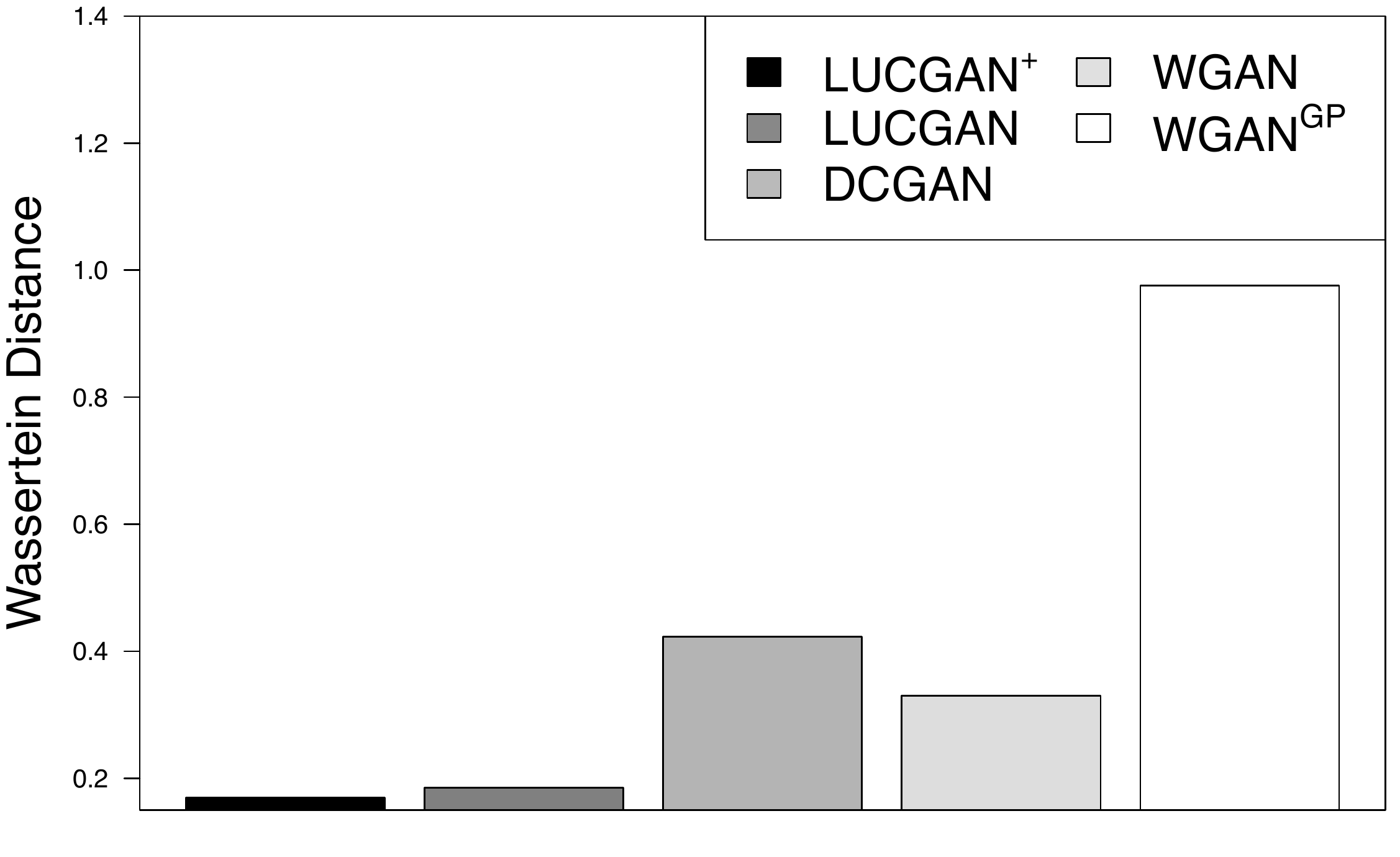}}}
	\caption{Overall performance for land-use configuration generation.}
	\label{fig:overall_performance}
	\vspace{-0.3cm}
\end{figure*}

To validate the effectiveness of our model, we evaluate the gap between the distribution of well-planned configurations and the distribution of generated configurations in terms of KL Divergence (KL), JS Divergence (JS), Hellinger Distance (HD), and Wasserstein Distance (WD).
As Figure \ref{fig:overall_performance} shows, compared with the best performance of baseline models (WGAN, WGAN$^{GP}$,DCGAN), LUCGAN$^{+}$ improves 16.2$\%$, 0.25$\%$, 28.4$\%$, 48.6$\%$ in terms of KL, JS, HD, and, WD respectively.
This observation indicates that LUCGAN$^+$ can capture more characteristics of the well-planned configurations compared with other baseline models.
In addition, another interesting observation is that compared with LUCGAN, 
LUCGAN$^+$ increases 8.92$\%$, 0.23$\%$, 8.43$\%$, 4.32$\%$ in terms of KL, JS, HD, and WD respectively. 
A potential interpretation for the observation is that the conditioning augmentation module and the new training approach of LUCGAN$^+$ makes the learning process more stable and effective.

\subsection{Study the influence of the square size for generating land-use configurations (Q2)}

\begin{figure*}[!thb]
\vspace{-0.3cm}
\setlength{\abovecaptionskip}{-2pt} 
	\centering
	\subfigure[KL Divergence]{\label{fig:robust_kl}\includegraphics[width=0.4\linewidth]{{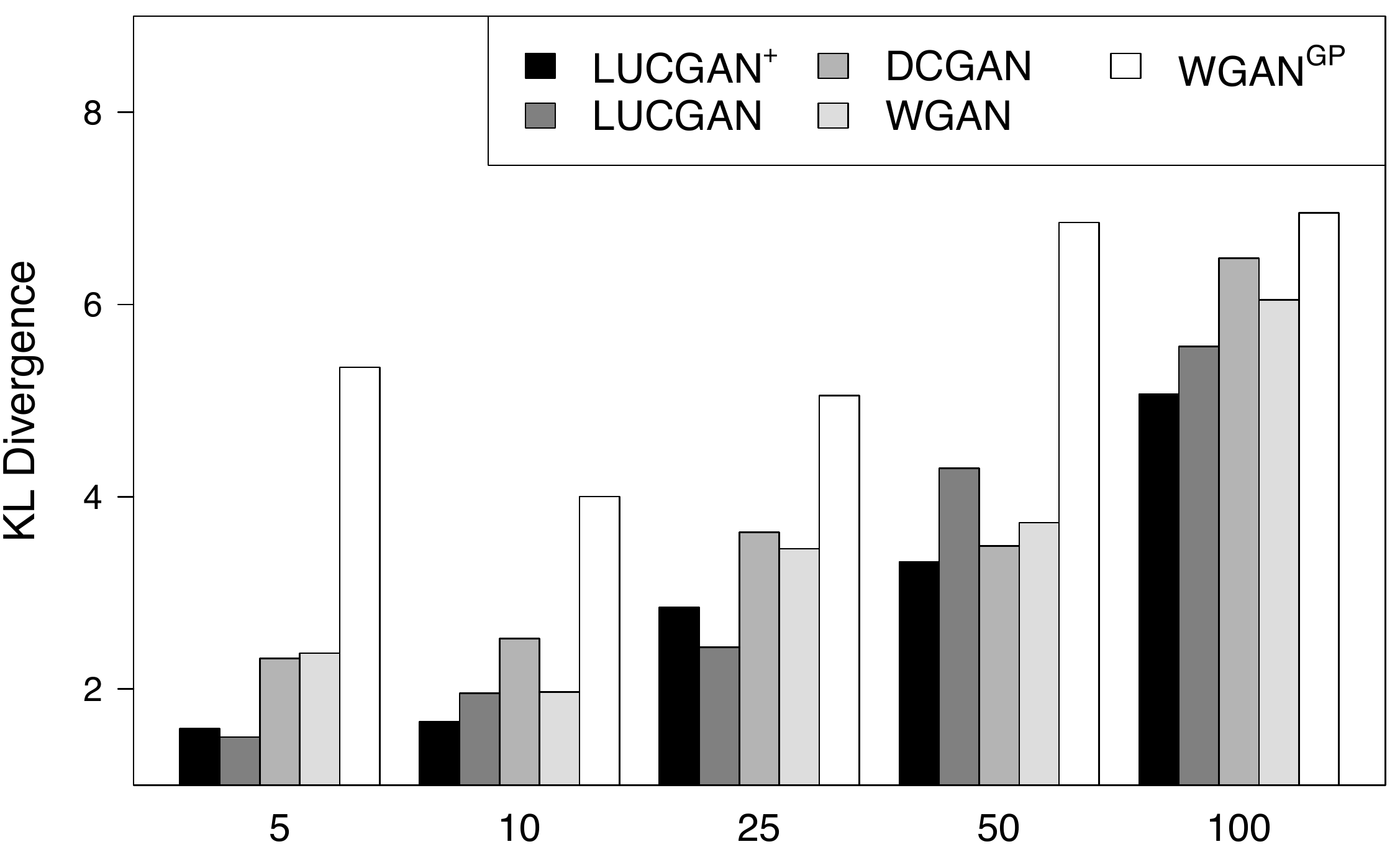}}}
	\subfigure[JS Divergence]{\label{fig:robust_js}\includegraphics[width=0.4\linewidth]{{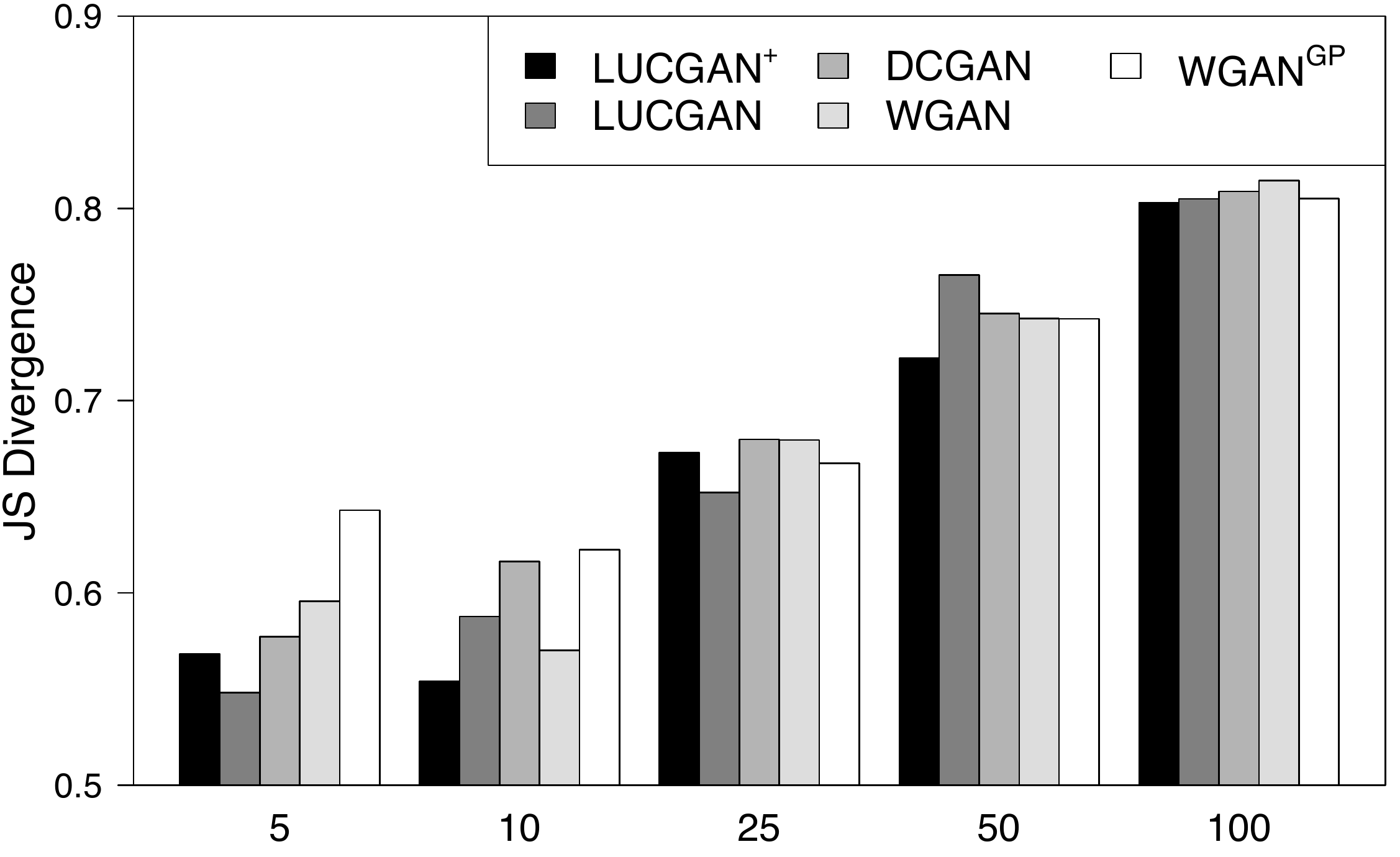}}}
	
	\subfigure[Hellinger Distance]{\label{fig:robust_hd}\includegraphics[width=0.4\linewidth]{{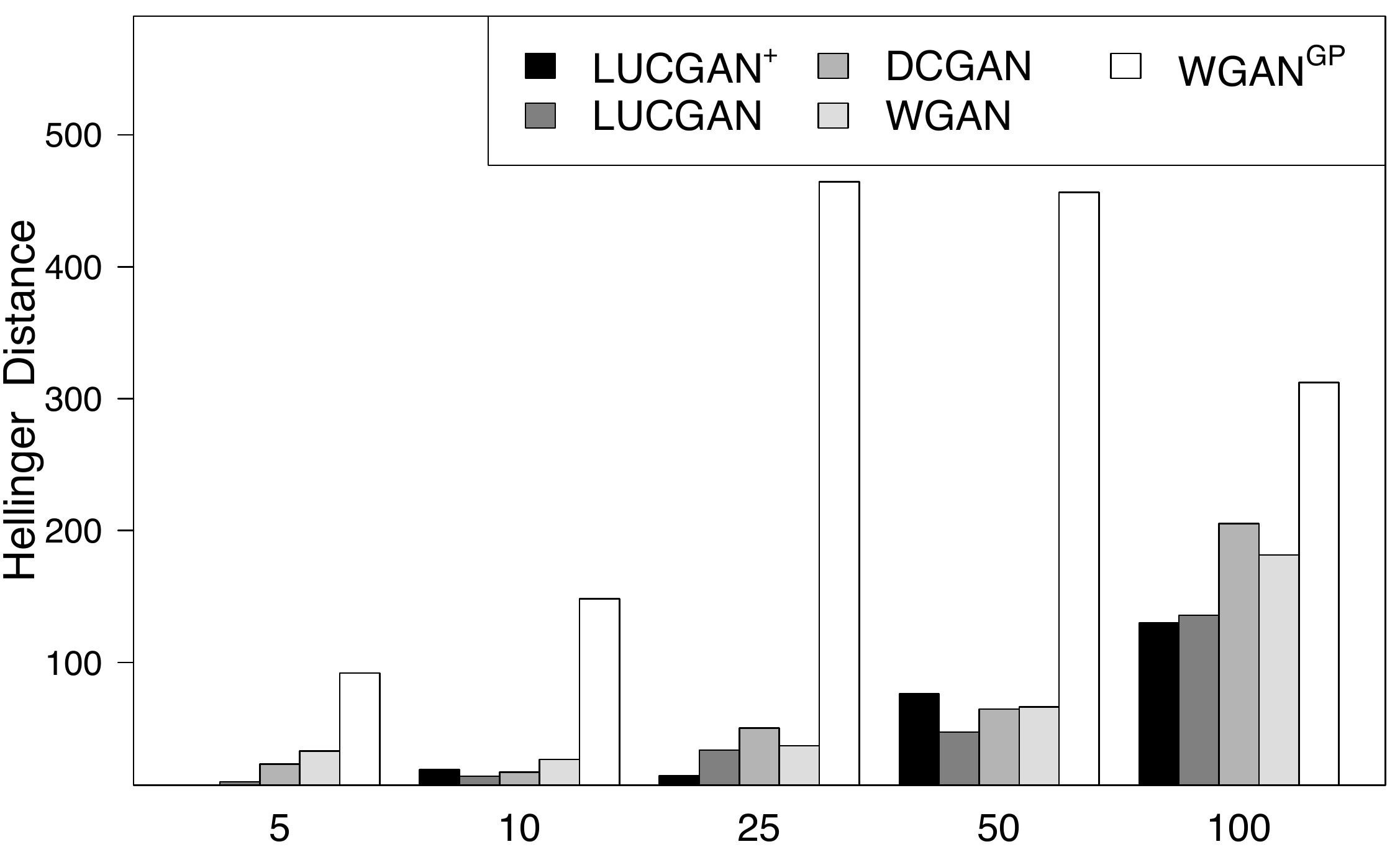}}}
	\subfigure[Wasserstein Distance]{\label{fig:robust_wd}\includegraphics[width=0.4\linewidth]{{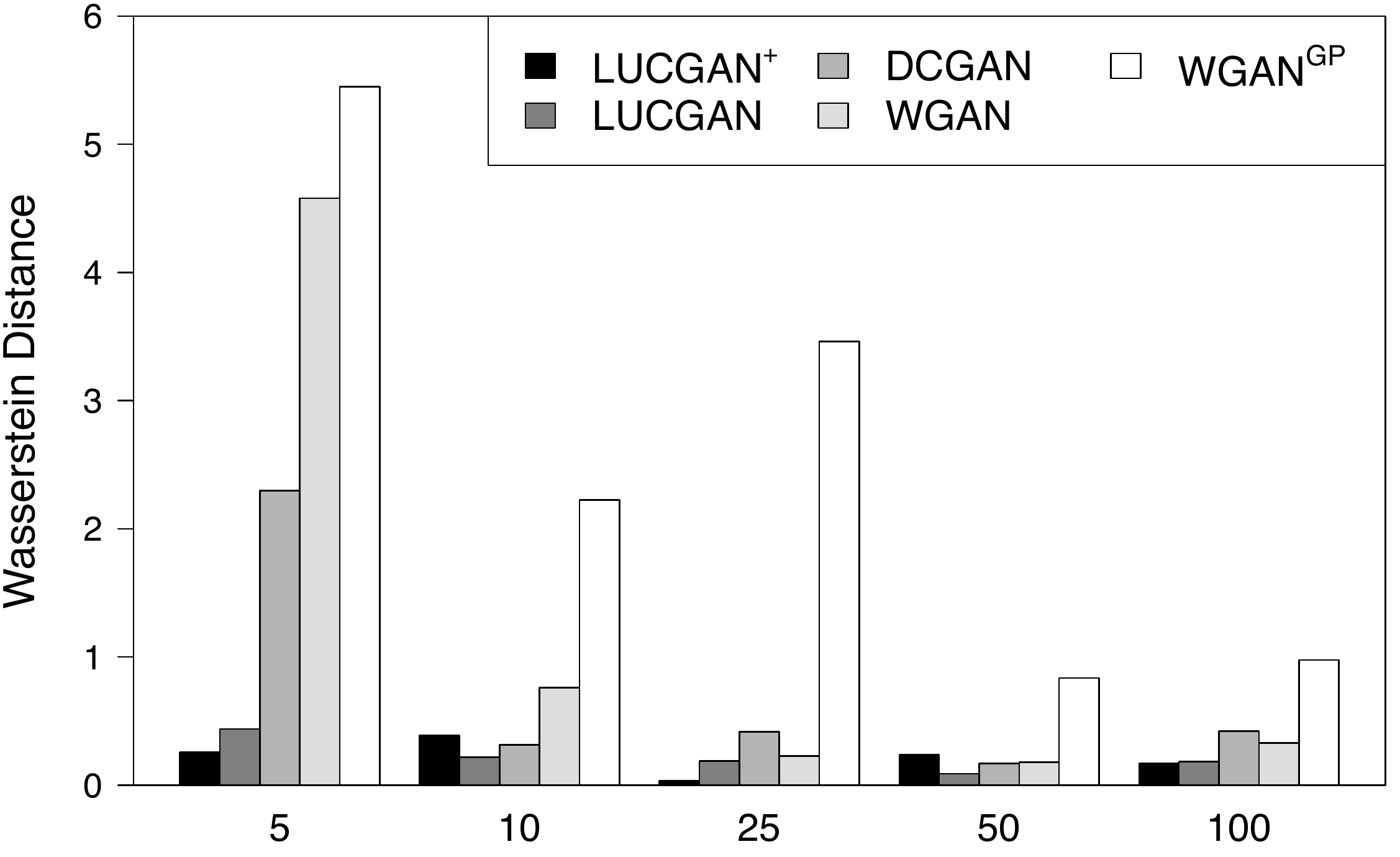}}}
	\caption{The influence of square size for the generation of land-use configurations.}
	\label{fig:robust_check}
	\vspace{-0.6cm}
\end{figure*}

To quantify the land-use configuration, we divide an area into $n\times n$ squares to collect the POI distribution information.
To study the influence of the square size for generation, we vary $n=5$, $n=10$, $n=25$, $n=50$, $n=100$ to conduct experiments.
Here, the smaller value of $n$ is, the larger size of square is.
Figure \ref{fig:robust_check} shows the performance of all models when facing different square sizes in terms of  KL Divergence, JS Divergence, HD, and WD.
We find that with the increase of the square size, the value of all metrics decreases. 
A possible explanation for the observation is that when the square size is larger, the distribution of the land-use configurations becomes simpler.
The generative models can capture the characteristics of the distribution of the configurations very easily, thus, the values of all metrics become smaller. 
However, the large square size loses much information about urban planning details of the land-use configuration.
Another interesting observation is that LUCGAN$^+$ outperforms other baseline models in terms of all evaluation metrics when $n=100$. 
But for some smaller $n$ values, LUCGAN $^+$ is slightly worse than LUCGAN.
A potential reason for the observation is that LUCGAN is enough to capture the pattern of land-use configurations collected by smaller $n$ values.
Although the conditional augmentation module of LUCGAN$^+$ can improve robustness, in this situation such module 
may cause the model to a slightly underfitting.
However, in reality, we should avoid collecting land-use configurations under small $n$ values.
Because such configurations lose many planning details, which is harmful to producing effective urban plans.

\subsection{Study the surrounding contexts of different configurations (Q3)}
\begin{figure}[htbp]
\vspace{-0.3cm}
\centering
\includegraphics[width=0.5\linewidth]{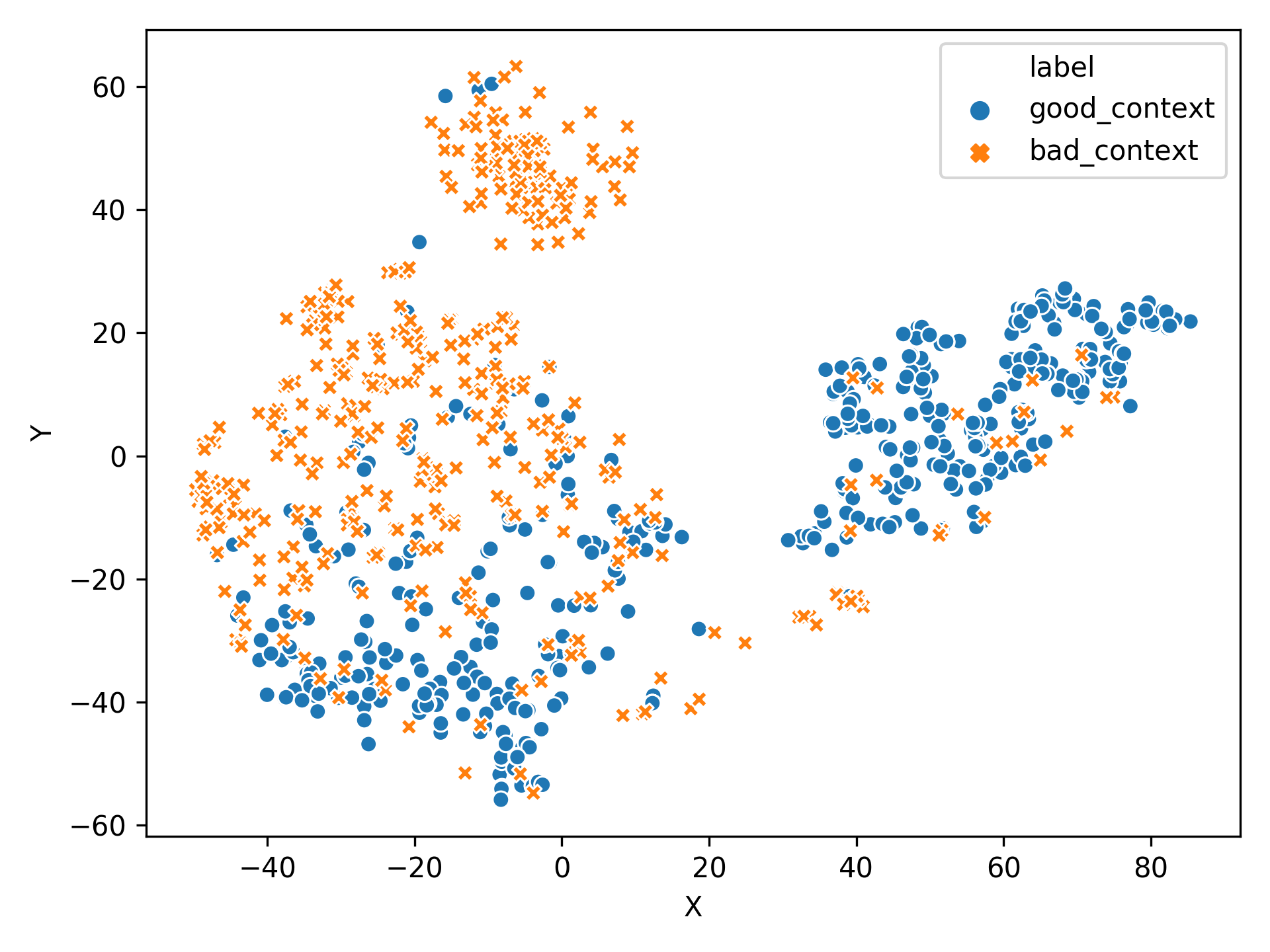}
\vspace{-0.6cm}
\caption{Visualization for the surrounding contexts of  well-planned and poorly-planned configurations.}
\label{emb_context}
\vspace{-0.55cm}
\end{figure}

Our framework generates land-use configurations based on the corresponding surrounding contexts.
Thus, the surrounding contexts have strong influence on the generation of the land-use configuration.
To observe the distribution of the surrounding contexts, we visualize the embeddings of the surrounding contexts on 2-dimensional space.
Specifically, we first randomly choose 500 embeddings of the surrounding context of well-planned configurations and poorly-planned configurations respectively.
Then, we utilize T-SNE algorithm ~\cite{van2008visualizing} to reduce the dimension of the embeddings into two.
Next, we visualize the embeddings on 2-dimensional space, as illustrated in Figure ~\ref{emb_context}.
We find that the pattern of the well-planned configurations contexts is different from the pattern of the poorly-planned configurations contexts, which indicates that our research intuition, that generates the land-use configurations based on the surrounding contexts is reasonable.

\subsection{Scoring model evaluation for generated land-use configurations (Q1)}
\begin{figure}[htbp]
\vspace{-0.3cm}
\centering
\includegraphics[width=0.5\linewidth]{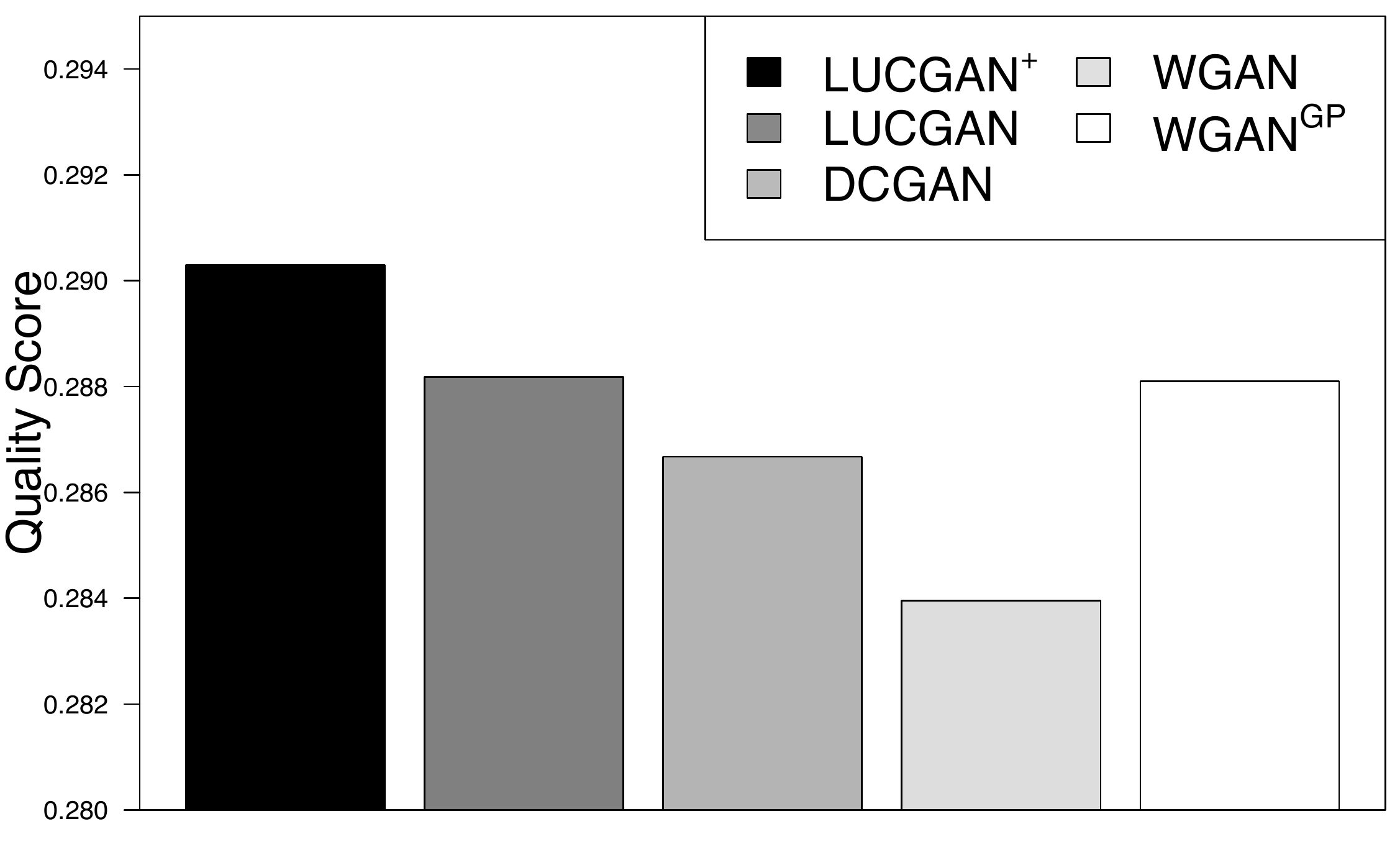}
\vspace{-0.3cm}
\caption{The quality score for different generated methods.}
\label{score_model}
\vspace{-0.6cm}
\end{figure}

% {\color{red}Delete??? Owing to urban planning is an open research area, there is no standard way to quantify and evaluate a land-use configuration.} 
To validate the effectiveness of LUCGAN$^+$ further, we build a scoring model.
As illustrated in Figure ~\ref{score_model}, LUCGAN$^+$ owns the highest quality score compared with other baseline models, which indicates the superiority of LUCGAN$^+$.
Meanwhile, it also shows that the scoring model can be regarded as a evaluation method for evaluating the generation of the land-use configurations.

\subsection{Study the POI ratio of generated configurations  under different $Q$ (Q4)}

\begin{figure*}[!thb]
\vspace{-0.5cm}
\setlength{\abovecaptionskip}{-2pt} 
	\centering
	\subfigure[Generated Vibrant Configuration]{\label{fig:vibrant_r}\includegraphics[width=0.45\linewidth]{{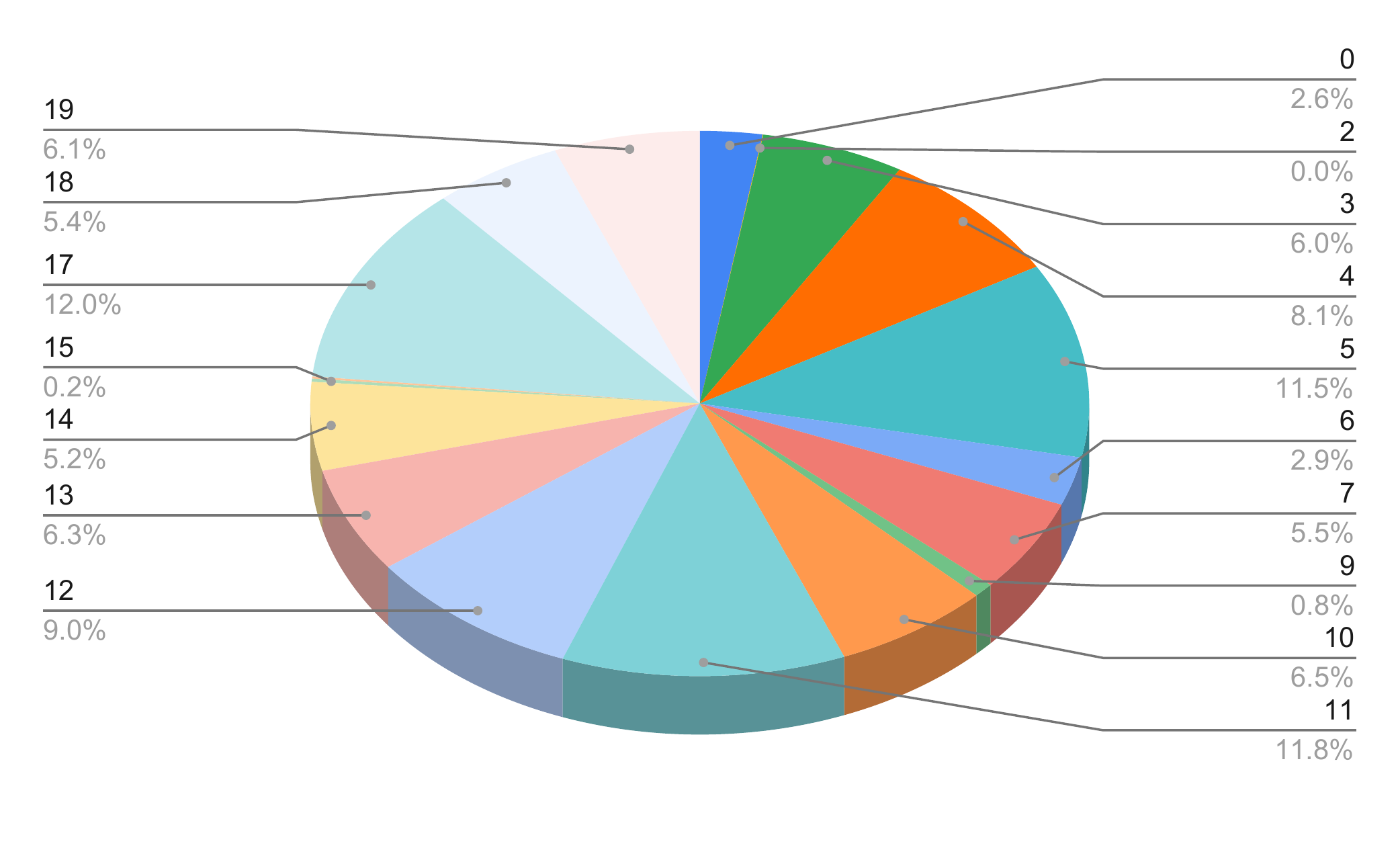}}}
	\subfigure[Original Vibrant Configuration]{\label{fig:vibrant_ro}\includegraphics[width=0.45\linewidth]{{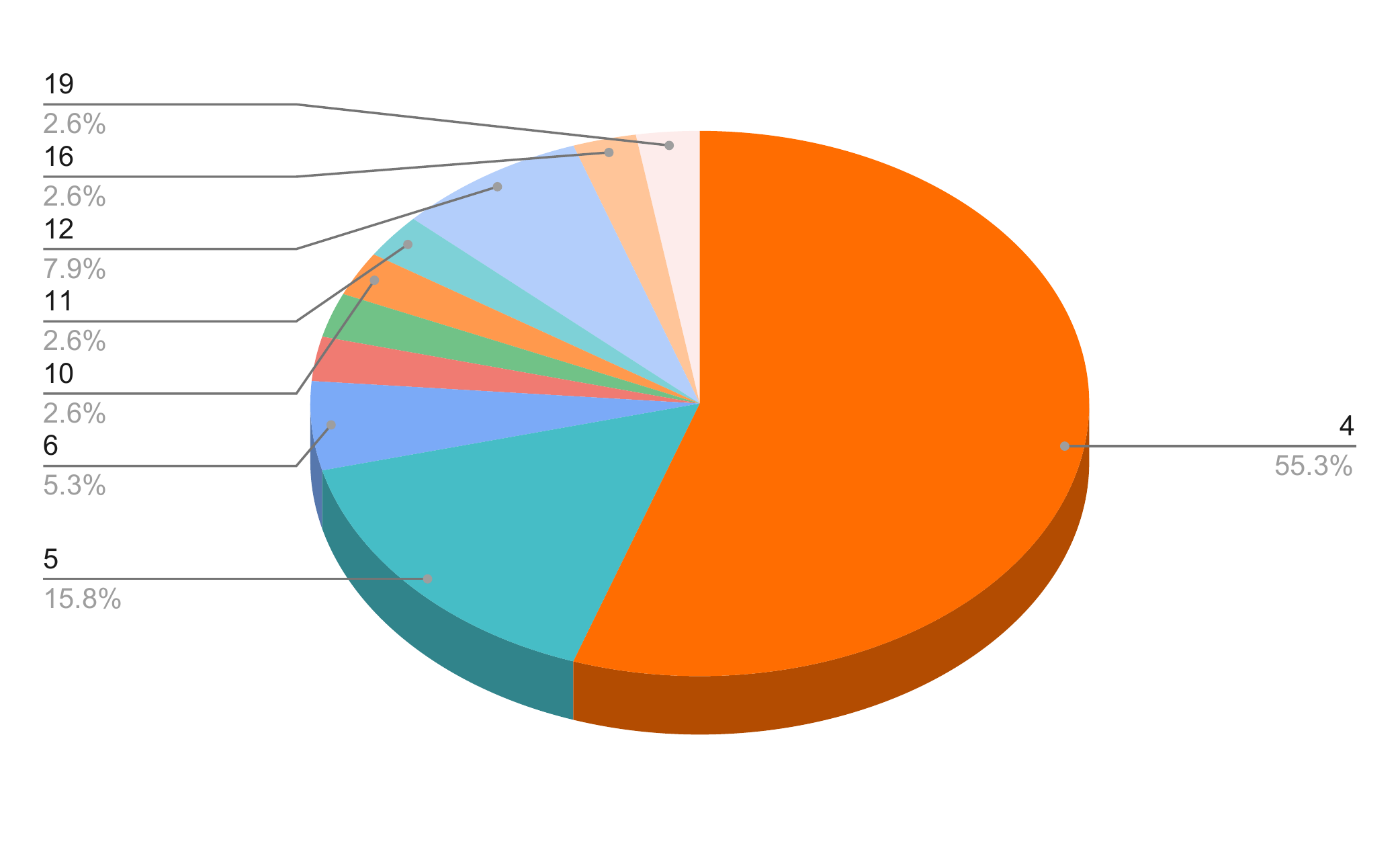}}}
	
	\subfigure[Generated Living Convenient Configuration]{\label{fig:living_r}\includegraphics[width=0.45\linewidth]{{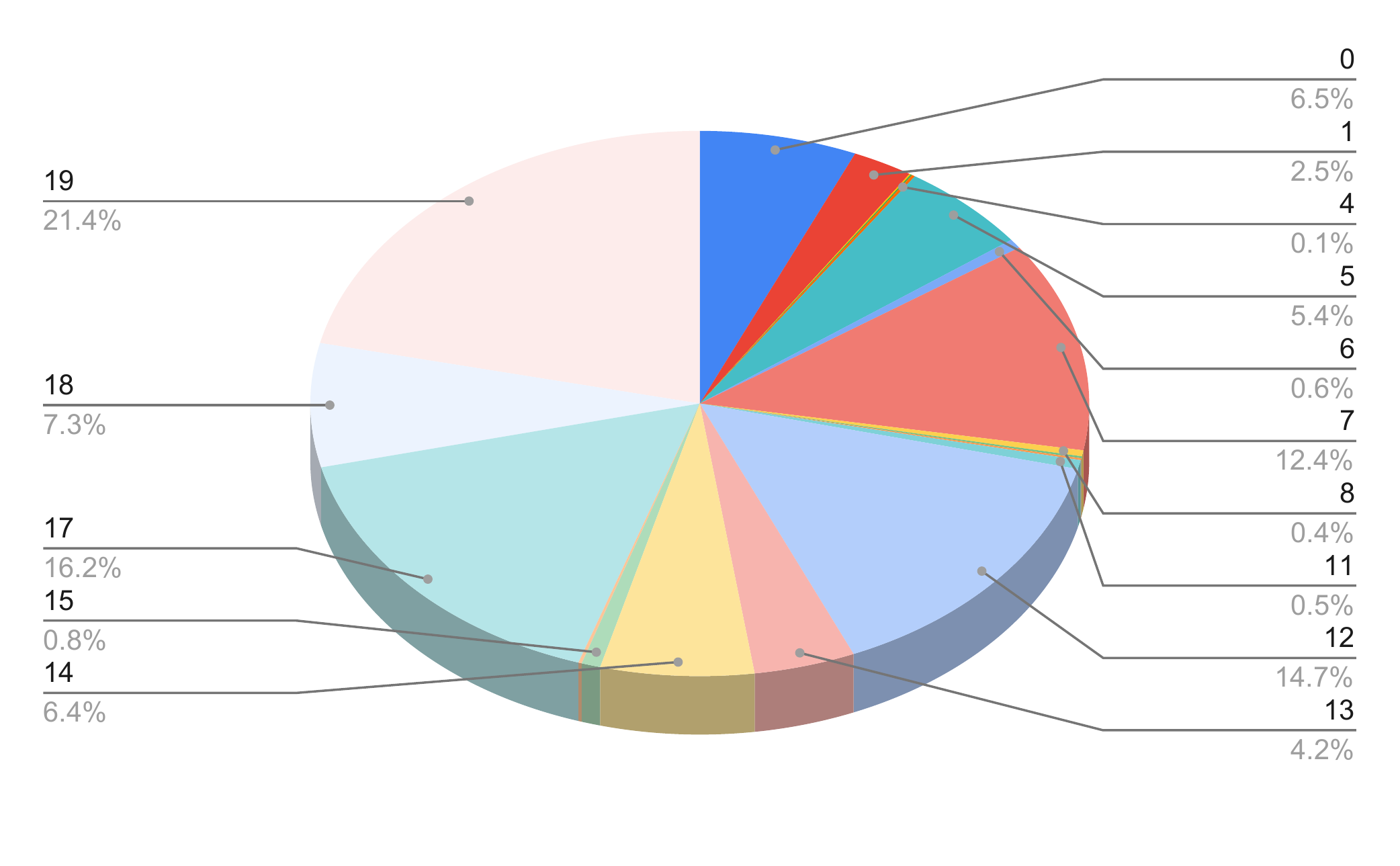}}}
	\subfigure[Original Living Convenient Configuration]{\label{fig:living_ro}\includegraphics[width=0.45\linewidth]{{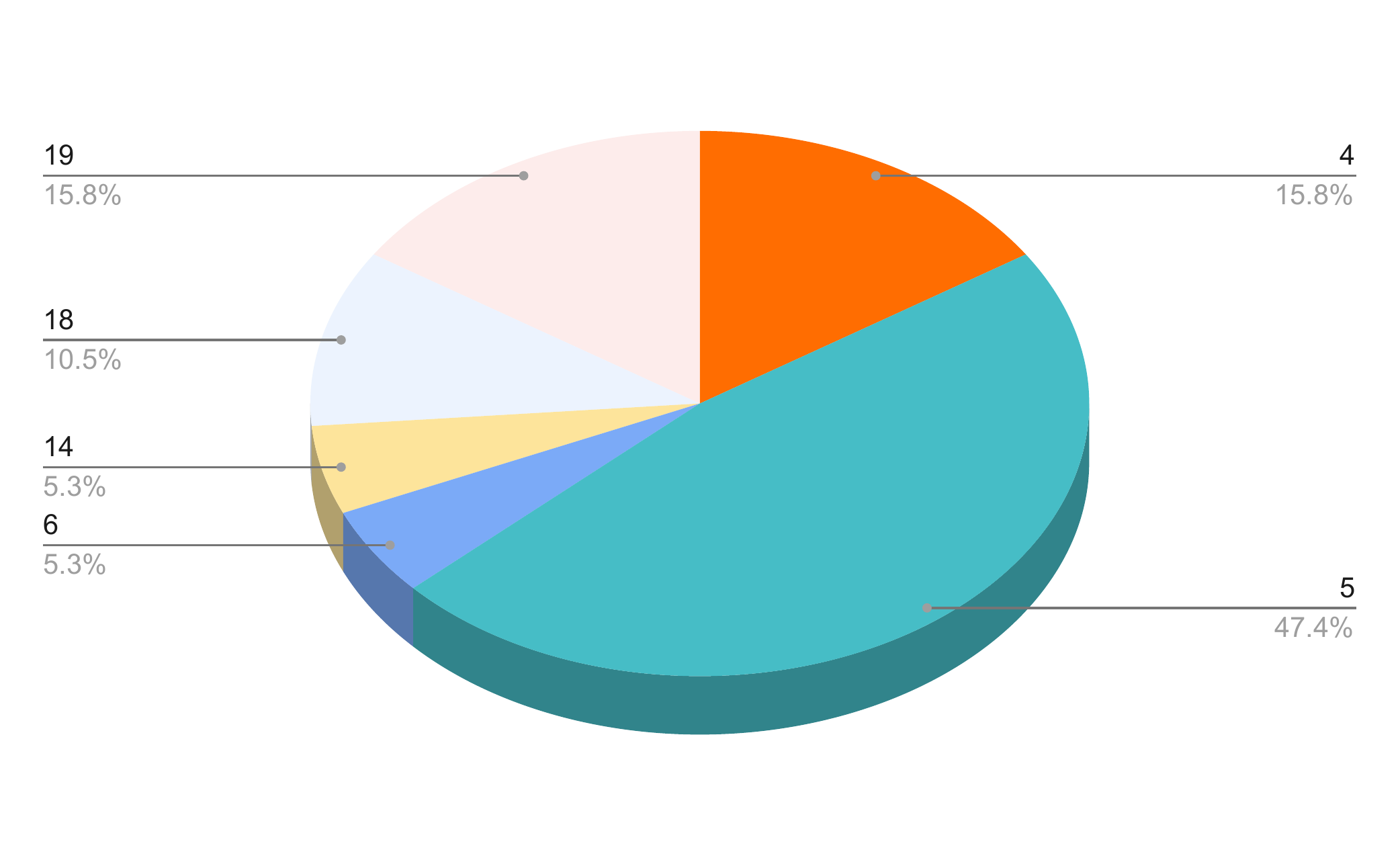}}}
	\caption{Comparison of POI ratio between generated and original configurations under different $Q$.}
	\label{fig:poi_ratio}
	\vspace{-0.6cm}
\end{figure*}

In our framework, we leverage a hyperparameter $Q$ to evaluate whether a land-use configuration is well-planned or poorly-planned.
When individuals have different urban planning goals, the meanings of $Q$ are different.
To validate the utility of $Q$, we conduct two generative tasks: (1) $Q$ is used to determine whether a land-use configuration is vibrant; (2) $Q$ is used to validate whether a land-use configuration is living convenient.
We visualize the POI ratio of generated configurations and original configurations under the two $Q$ settings as shown in Figure ~\ref{fig:poi_ratio}, in which the numbers $0 \sim 19$ denote different POI categories that are shown Table ~\ref{poi_lists}, and the grey percentiles indicate the proportions of different POI categories in the corresponding configuration.
Compared with Figure ~\ref{fig:vibrant_r} and Figure ~\ref{fig:living_r}, we find that for the vibrant configuration, POI category 4 (food service), 5 (shopping), 7 (recreation service), and 11 (real estate) cover a large portion.
This is reasonable because a vibrant configuration always owns many POIs related to economics and social activities;
For the living convenient configuration, POI category 12 (government place), 17 (road furniture), and 19 (public service) occupy the majority.
A reasonable explanation is that a configuration is living convenient when it contains many POIs related to public services and traffic conditions. 
The two observations validate that LUCGAN$^+$ can produce the customized land-use configuration utilizing $Q$ according to people's requirements.
In addition, compared with Figure ~\ref{fig:vibrant_r} and Figure ~\ref{fig:vibrant_ro}, Figure ~\ref{fig:living_r} and Figure ~\ref{fig:living_ro}, another interesting observation is that the POI categories in generated configurations are more complete than the original configurations.
A potential interpretation is that LUCGAN$^+$ not only captures the characteristics of the specific kind of land-use configuration but also includes new design elements into the generated configuration.

\subsection{Study the POI distribution of generated configurations under different $Q$ (Q4)}

\begin{figure*}[!thb]
\vspace{-0.5cm}
\setlength{\abovecaptionskip}{-2pt} 
	\centering
	\subfigure[Generated Vibrant Configuration]{\label{fig:vibrant_p}\includegraphics[width=0.45\linewidth]{{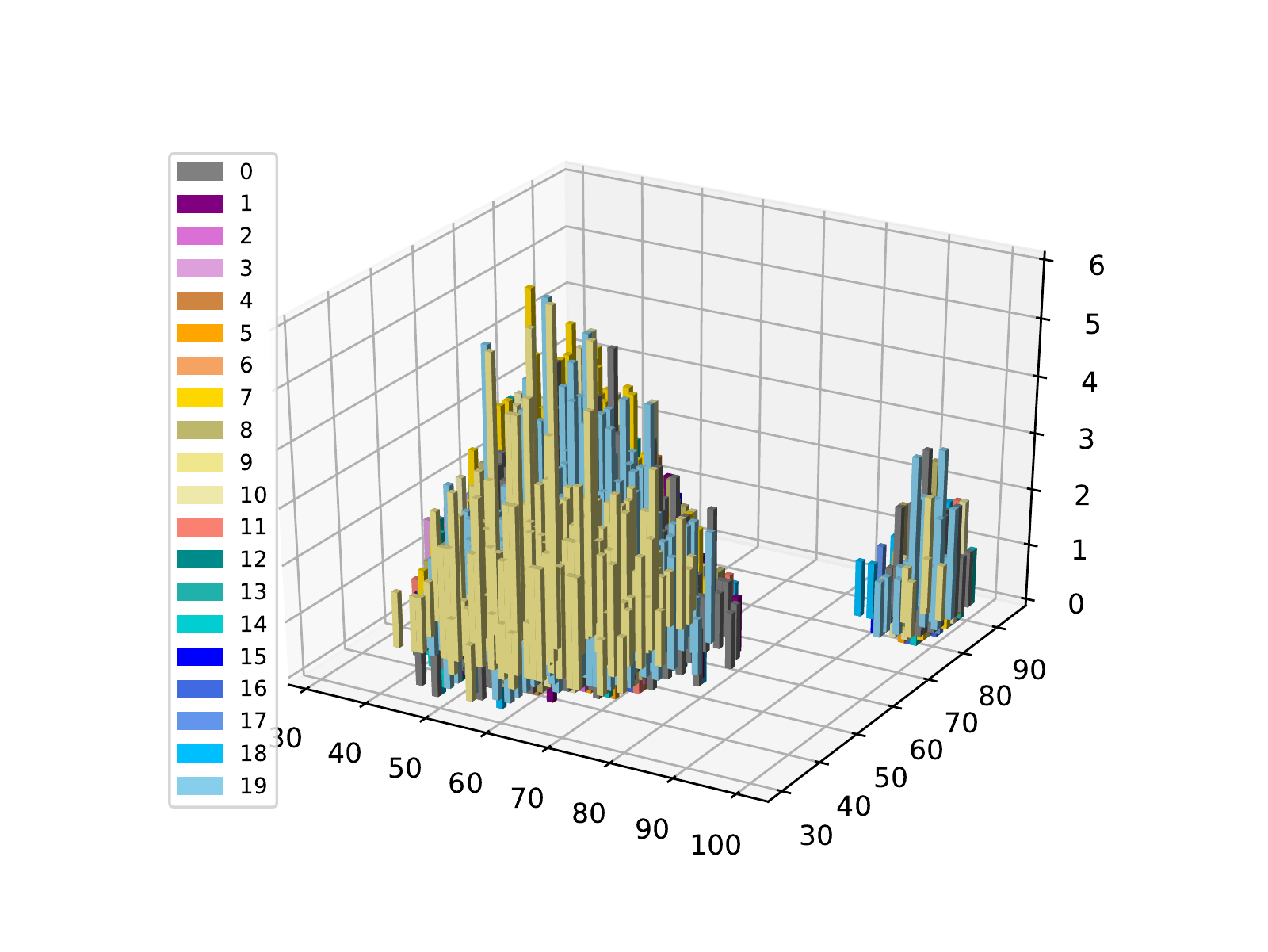}}}
	\subfigure[Original Vibrant Configuration]{\label{fig:vibrant_po}\includegraphics[width=0.45\linewidth]{{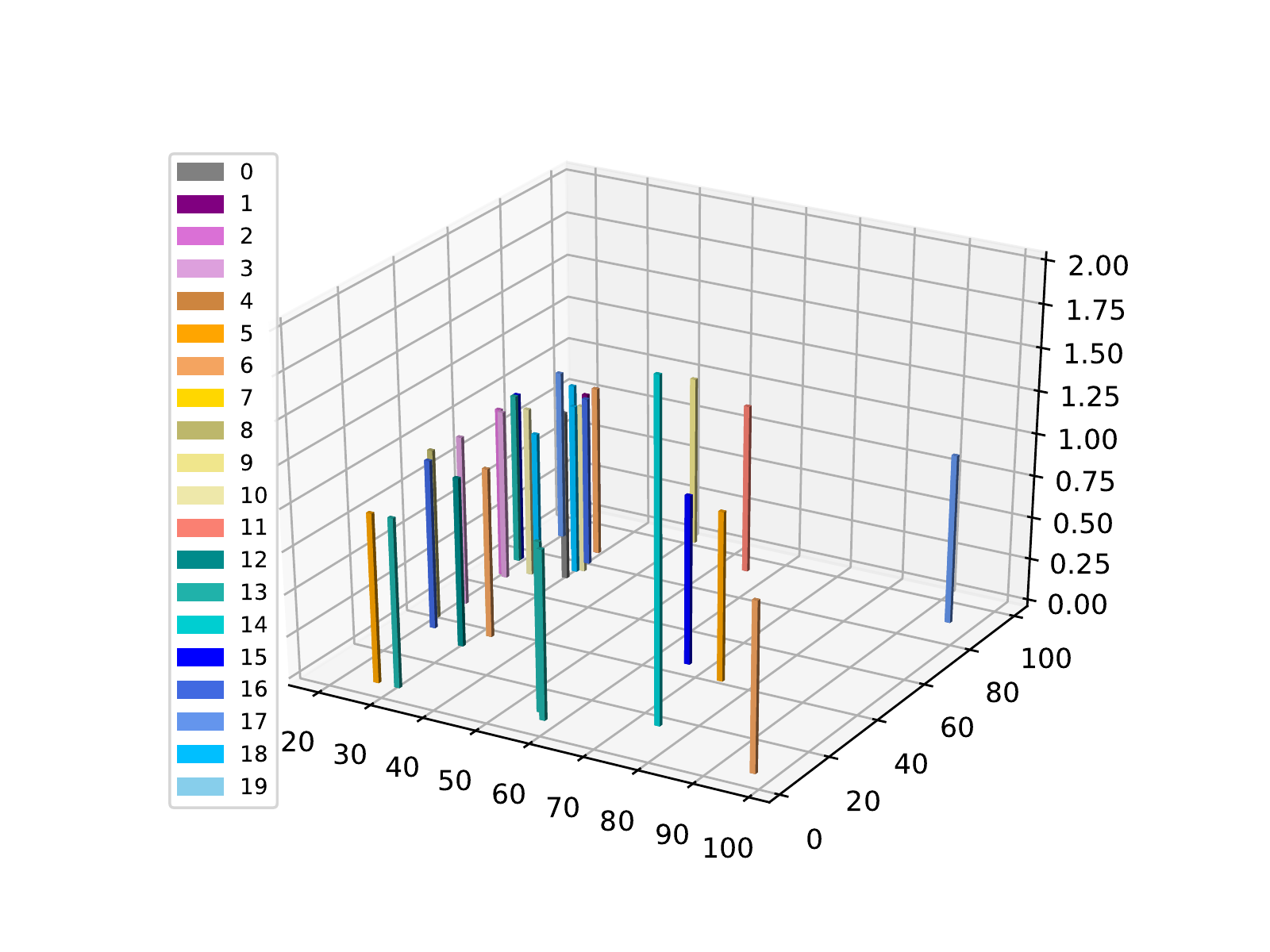}}}
	
	\subfigure[Generated Living Convenient Configuration]{\label{fig:living_p}\includegraphics[width=0.45\linewidth]{{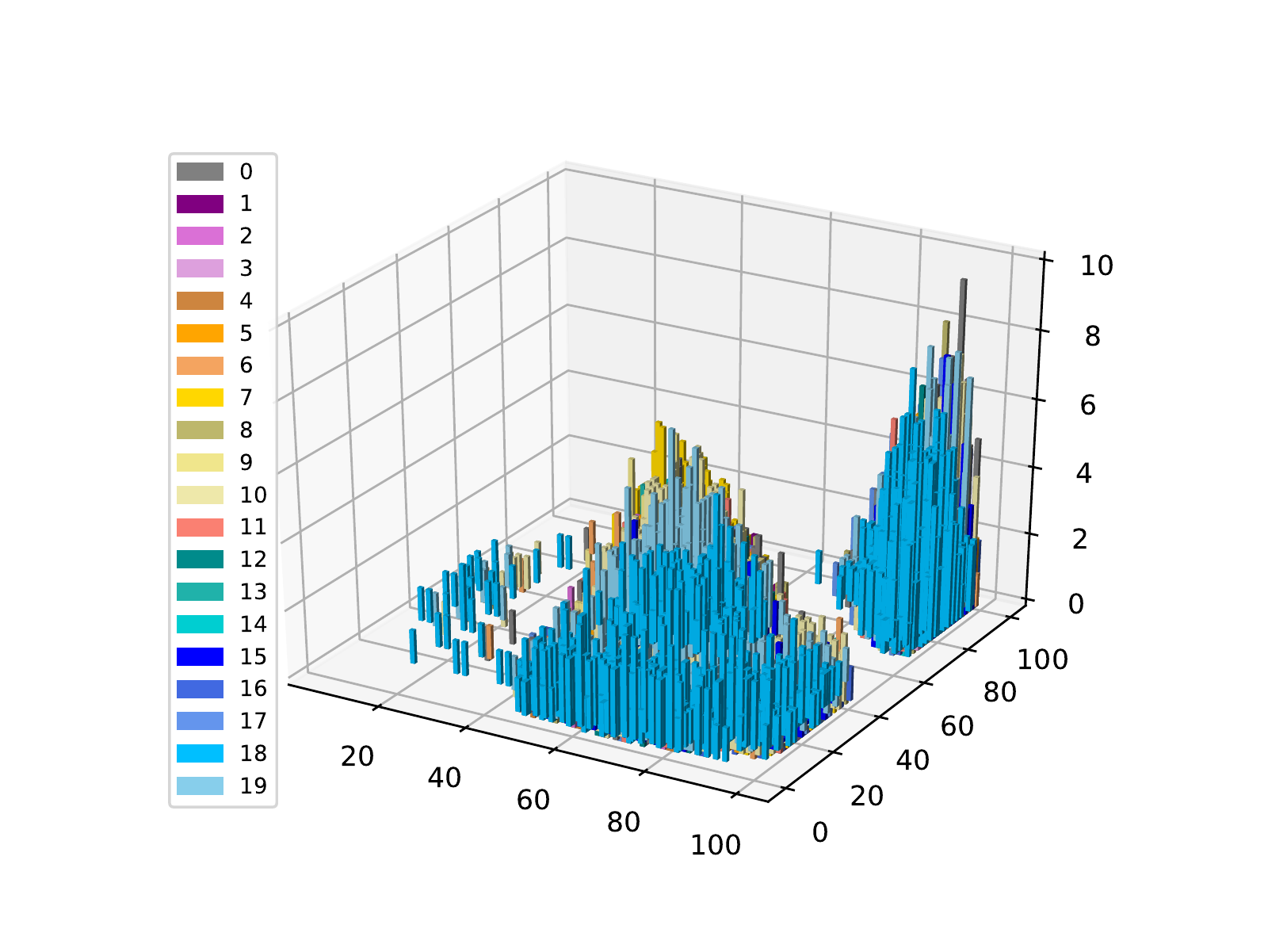}}}
	\subfigure[Original Living Convenient Configuration]{\label{fig:living_po}\includegraphics[width=0.45\linewidth]{{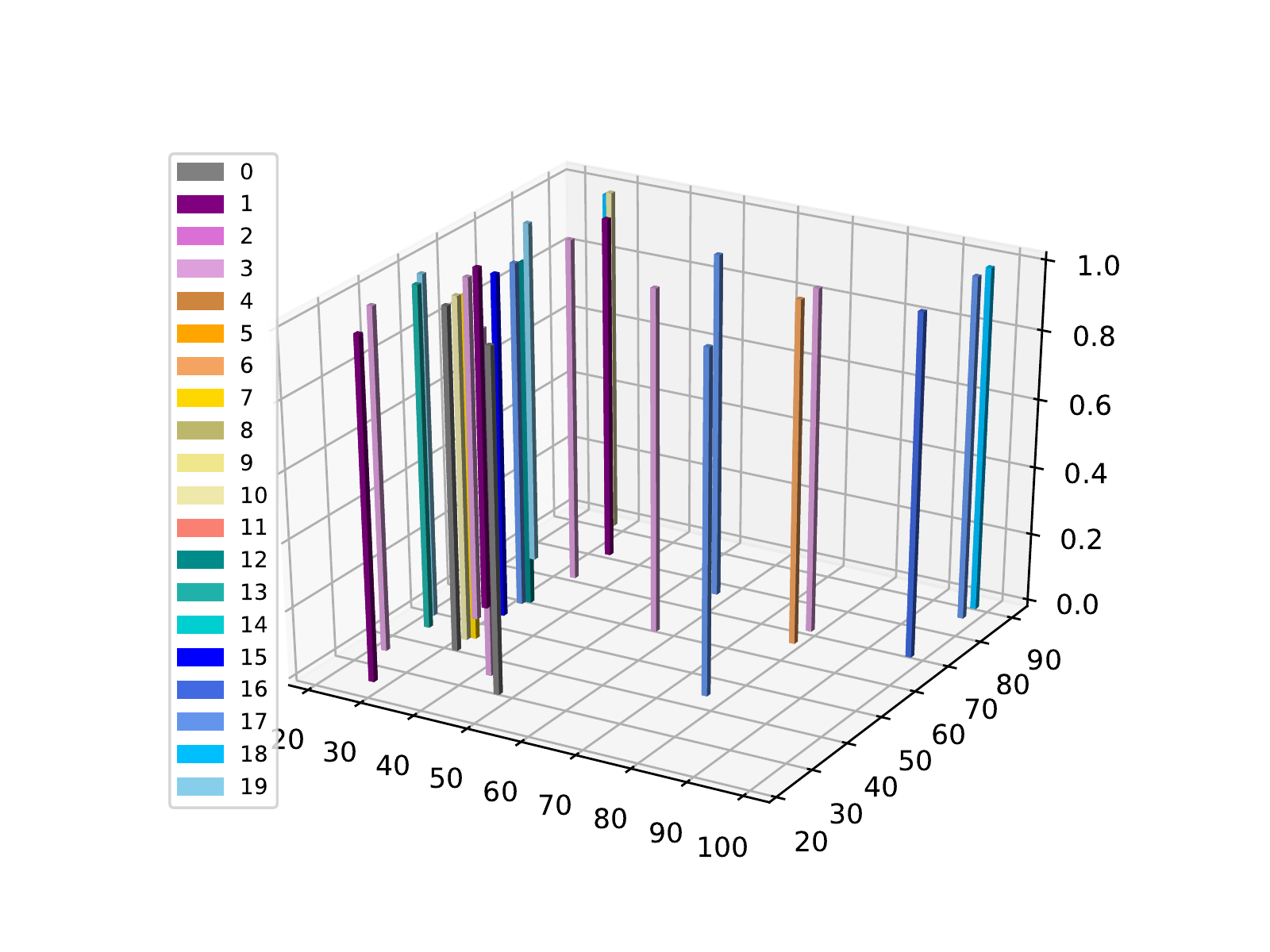}}}
	\caption{Comparison of POI distribution between generated and original configurations under different $Q$}
	\label{fig:poi_dis}
	\vspace{-0.6cm}
\end{figure*}

To further understand the utility of $Q$ and observe the differences between generated and original land-use configurations, we visualize the configurations into a 3-dimensional space as shown in Figure ~\ref{fig:poi_dis}, in which the left color bar indicates the mapping relations between the number of POI categories and colors; 
the right part reflects the POI distribution of the configuration; 
the height of each bar indicates the number of POIs at the corresponding position.
A careful inspection for Figure ~\ref{fig:vibrant_p} and Figure ~\ref{fig:living_p} shows that the generated configurations are organized and contain enough planning information for implementation in realistic.
In addition, another interesting observation is that the generated configurations contain more dense POI distribution compared with original configurations.
A potential interpretation is that LUCGAN$^+$ prefers to produce dense POI distribution, because it's easy to capture the correlation among different POIs.

% In addition, Figure ~\ref{fig:vibrant_poi} and Figure ~\ref{fig:living_poi} show the POI ratio of configurations.
% We find that for the vibrant configuration, POI category 4 (food service), 5 (shopping), 7 (recreation service), and 11 (real estate) cover a large part.
% This is reasonable because a vibrant configuration always owns many POIs related to economics and social activities;
% For the living convenient configuration, POI category 12 (government place), 17 (road furniture), and 19 (public service) occupy the majority.
% A reasonable explanation is that people live conveniently when the living configuration contains many POIs related to public services and traffic conditions.

\subsection{Study the generated situation of each channel in generated configurations (Q5)}

\begin{figure*}[!thb]
\vspace{-0.3cm}
\setlength{\abovecaptionskip}{-0pt} 
	\centering
	\subfigure[car service]{\label{fig:channel1}\includegraphics[width=0.24\linewidth]{{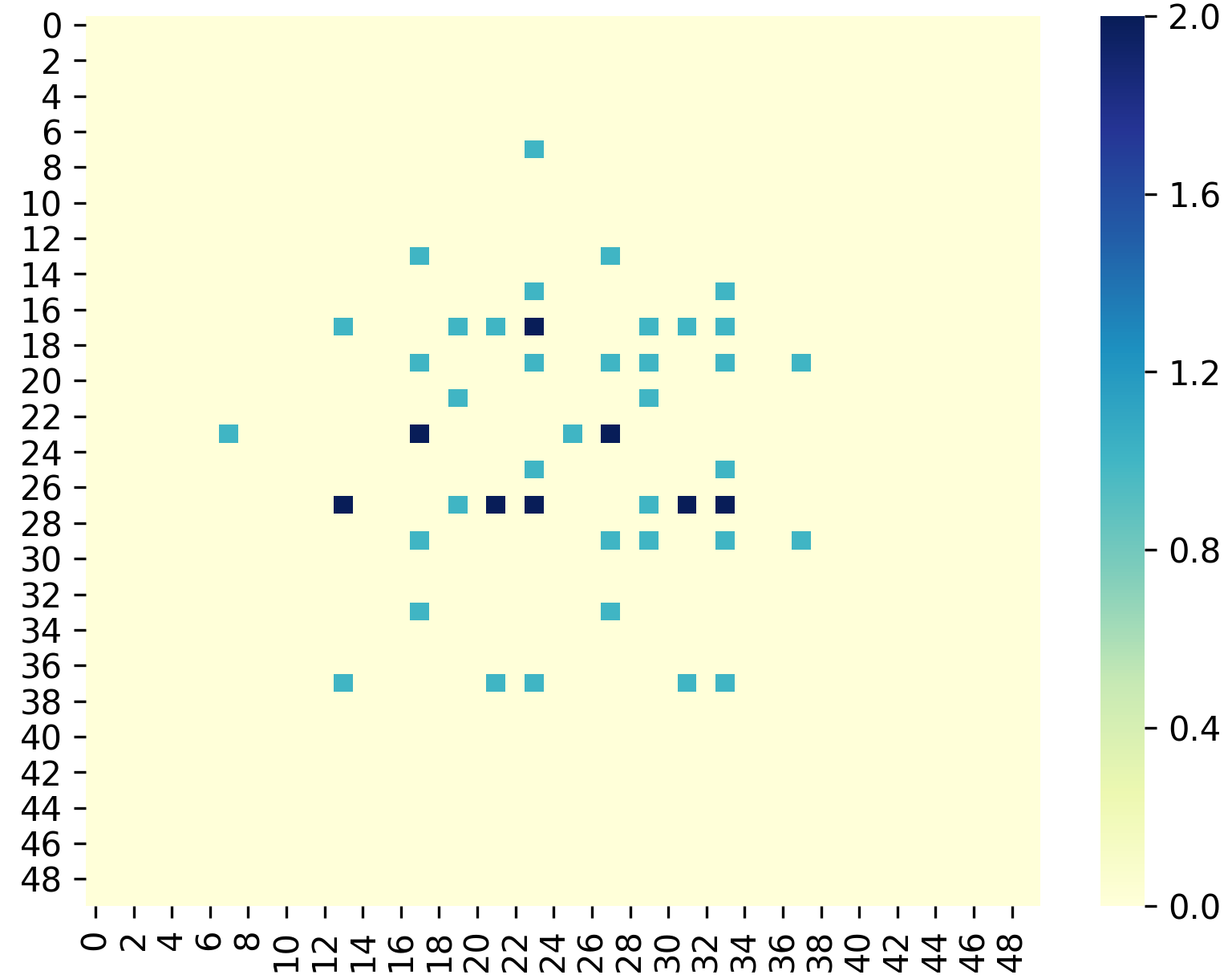}}}
	\subfigure[food service]{\label{fig:channel4}\includegraphics[width=0.24\linewidth]{{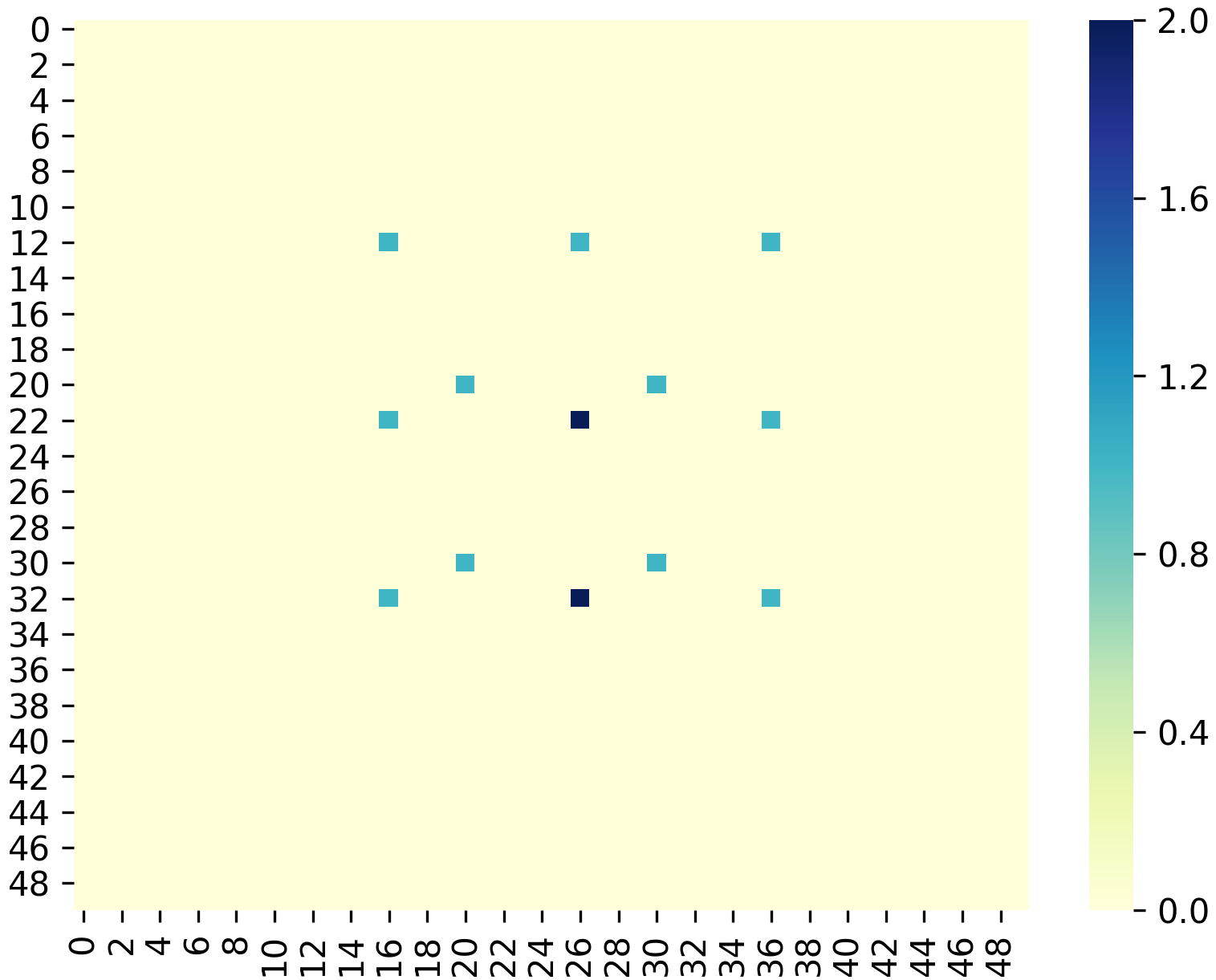}}}
	\subfigure[daily life service]{\label{fig:channel6}\includegraphics[width=0.24\linewidth]{{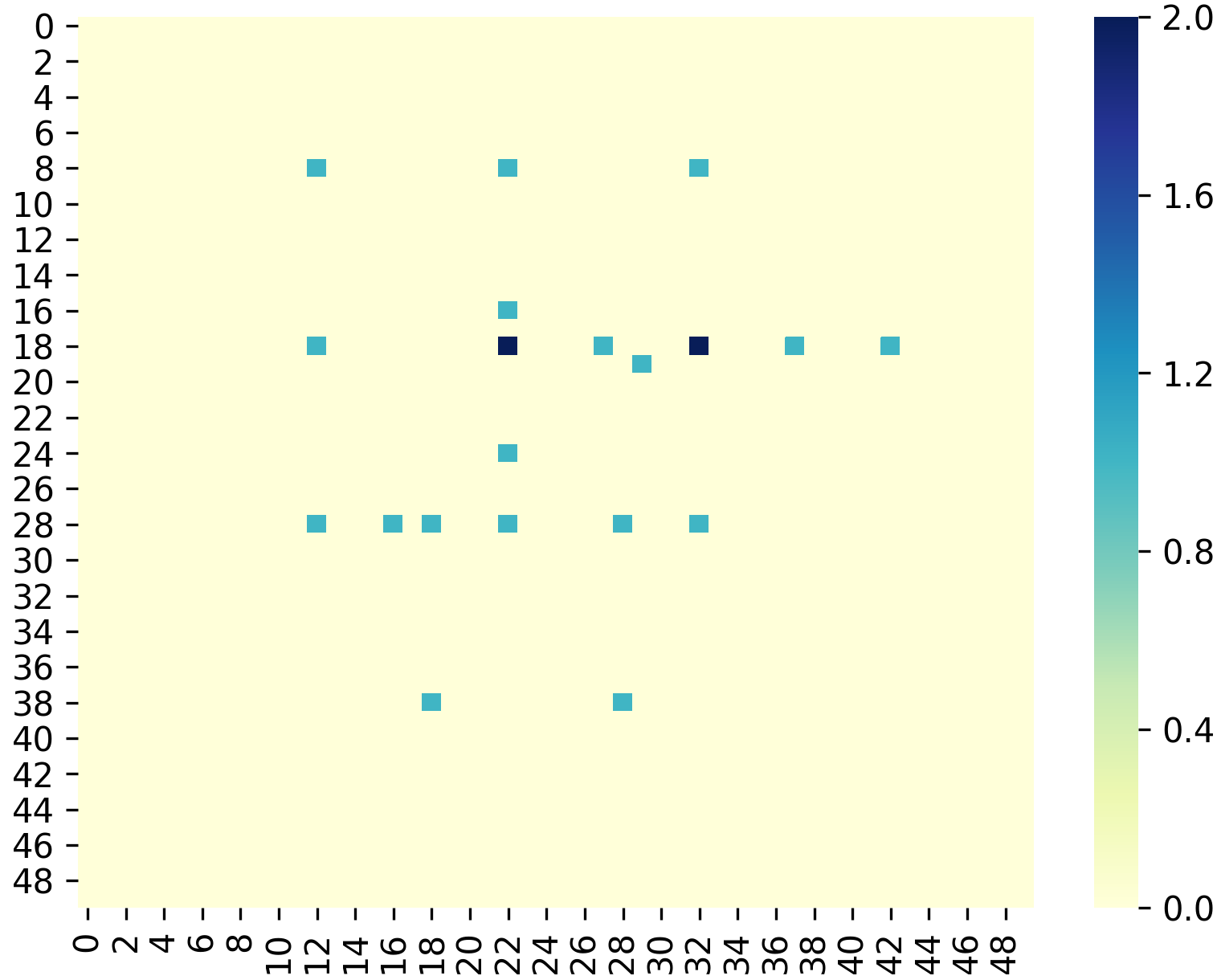}}}
	\subfigure[recreation service]{\label{fig:channel7}\includegraphics[width=0.24\linewidth]{{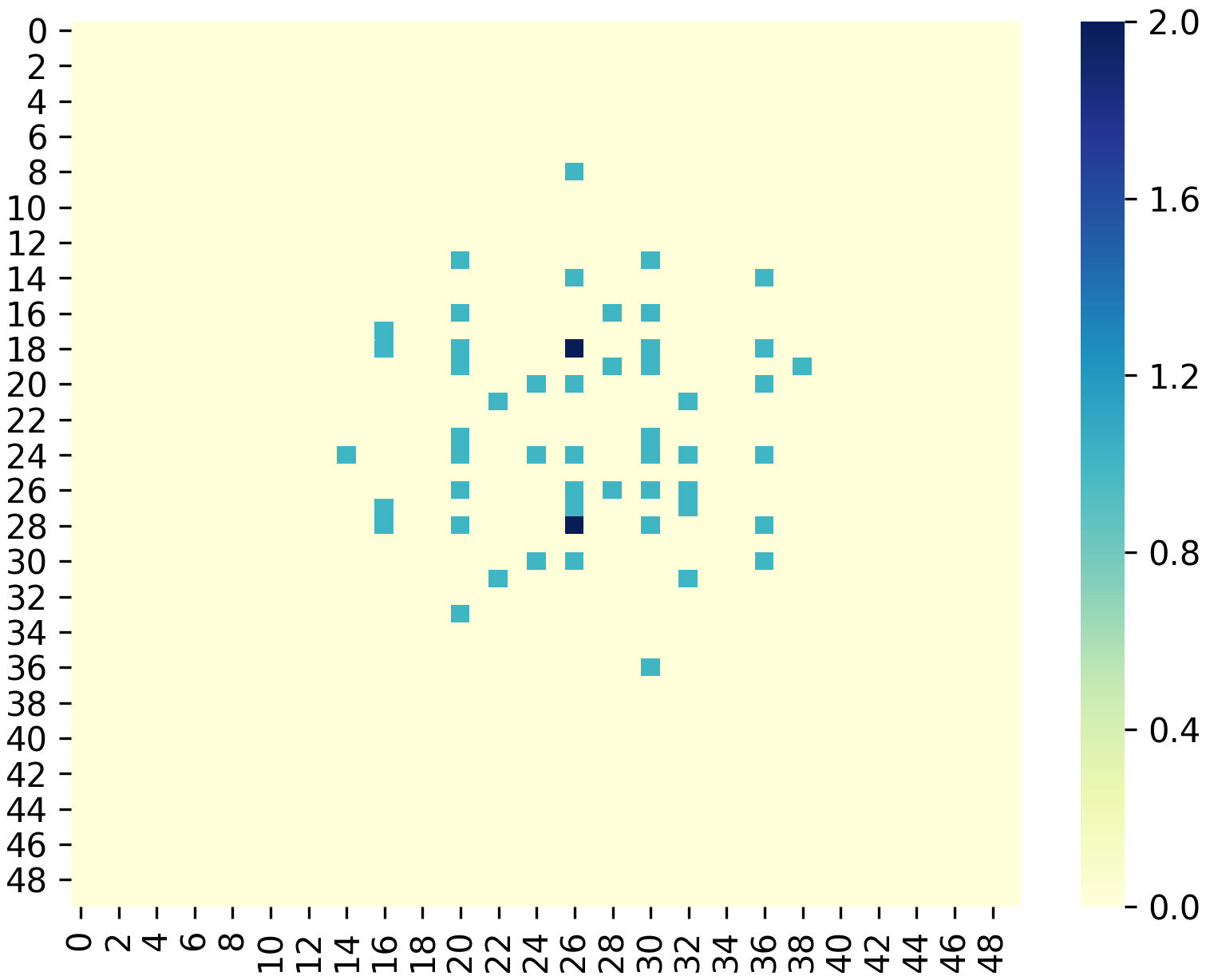}}}
	
	\vspace{-0.3cm}
	
	\subfigure[tourist attraction]{\label{fig:channel10}\includegraphics[width=0.24\linewidth]{{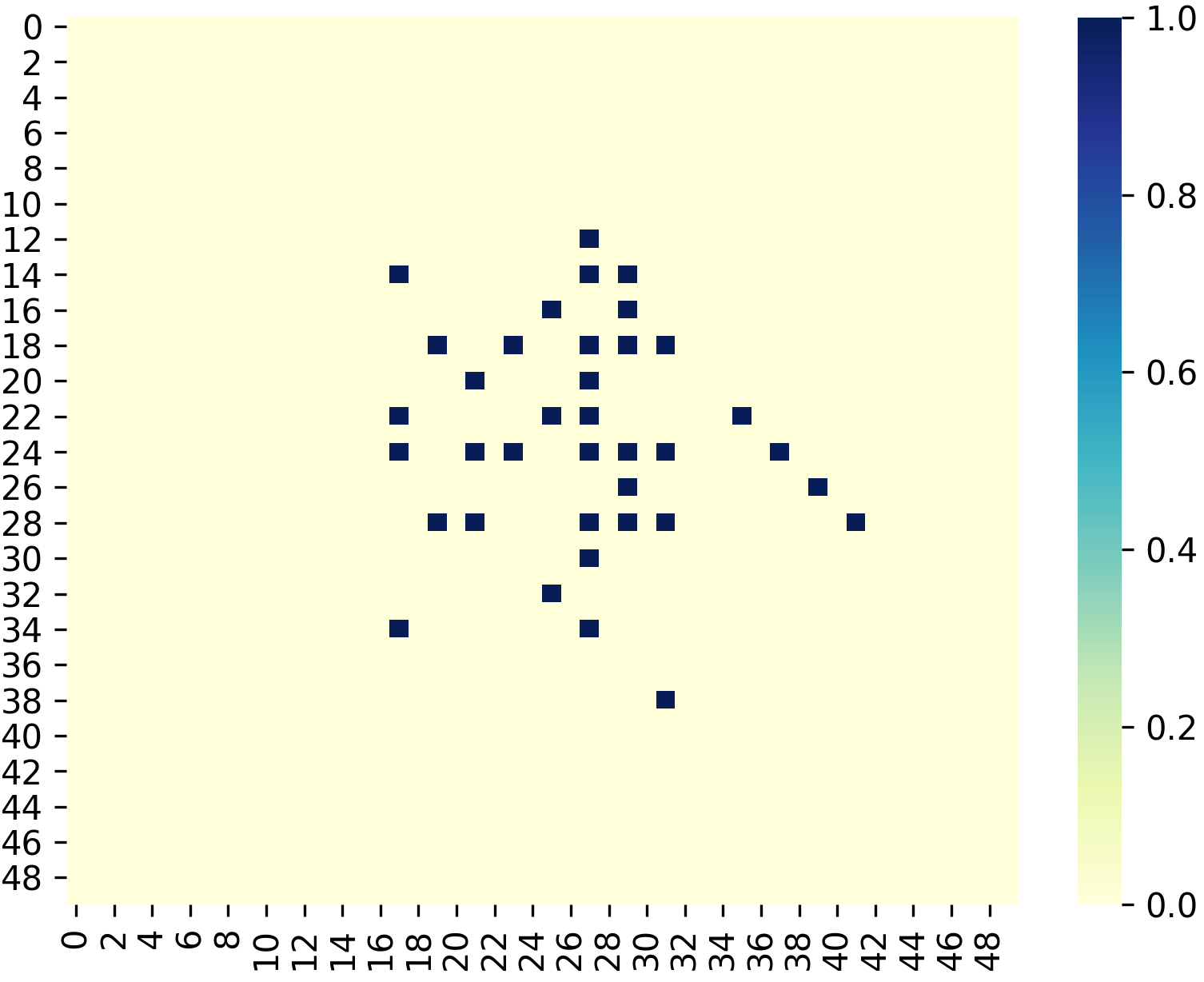}}}
	\subfigure[real estate]{\label{fig:channel11}\includegraphics[width=0.24\linewidth]{{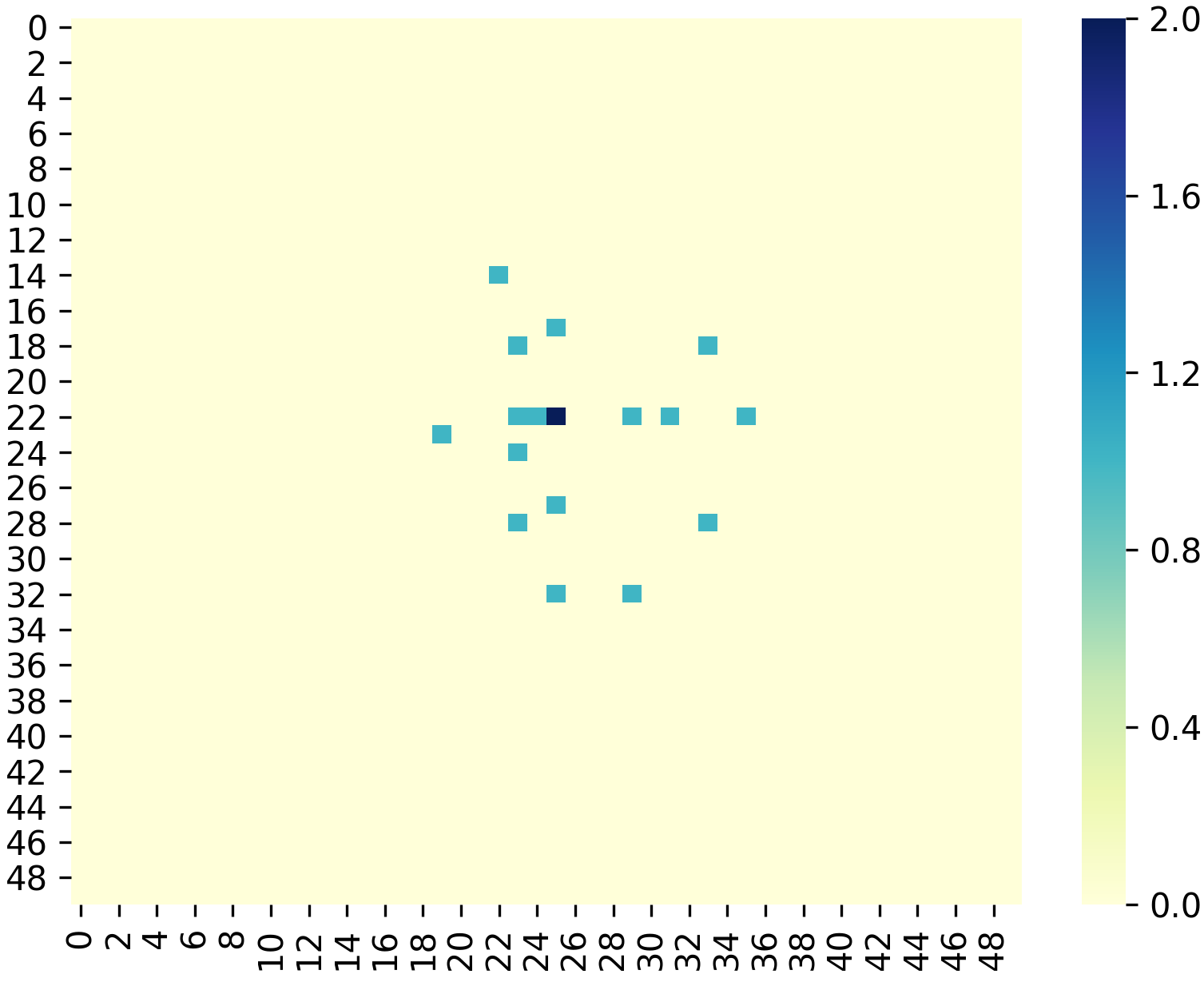}}}
	\subfigure[government place]{\label{fig:channel12}\includegraphics[width=0.24\linewidth]{{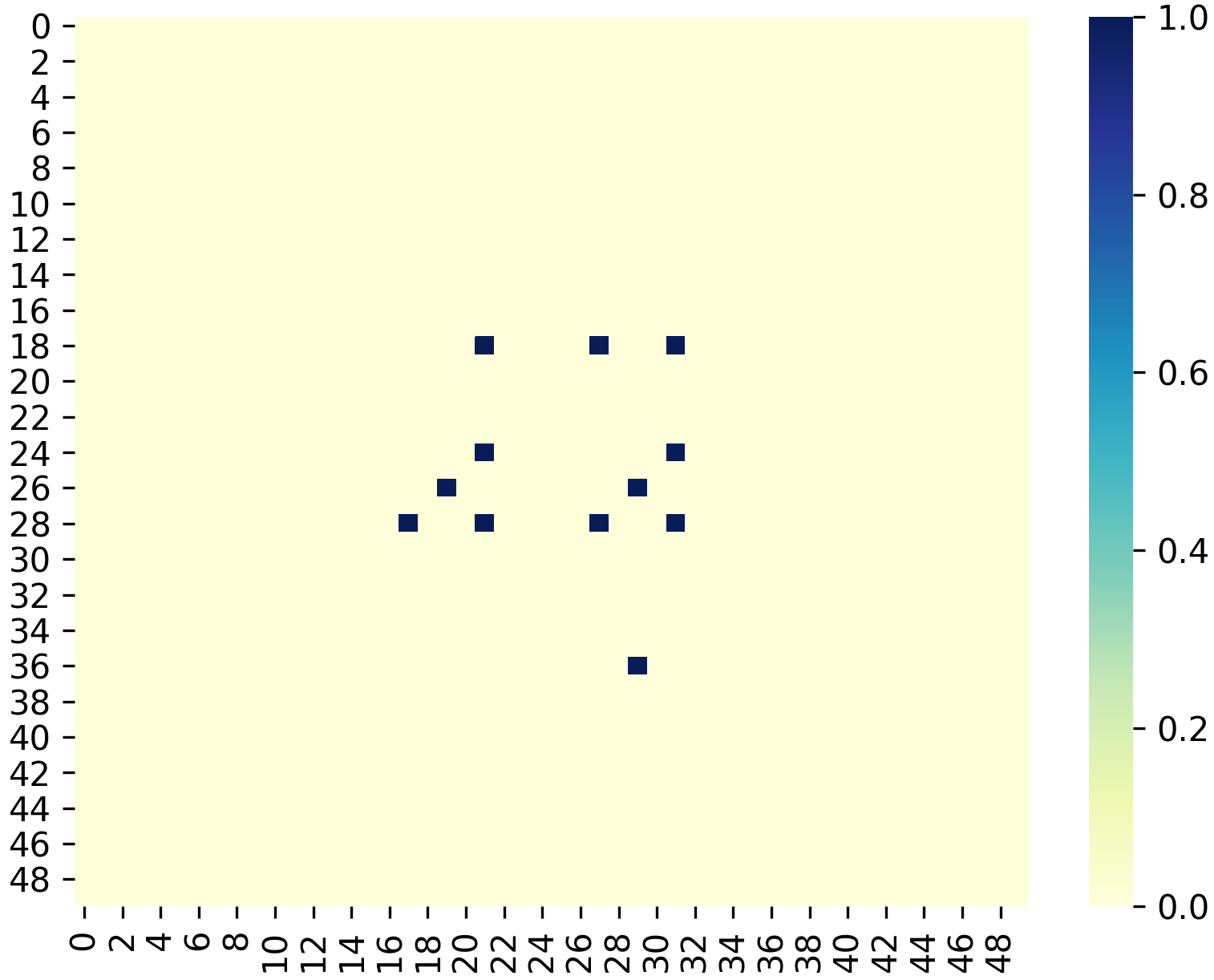}}}
	\subfigure[education]{\label{fig:channel13}\includegraphics[width=0.24\linewidth]{{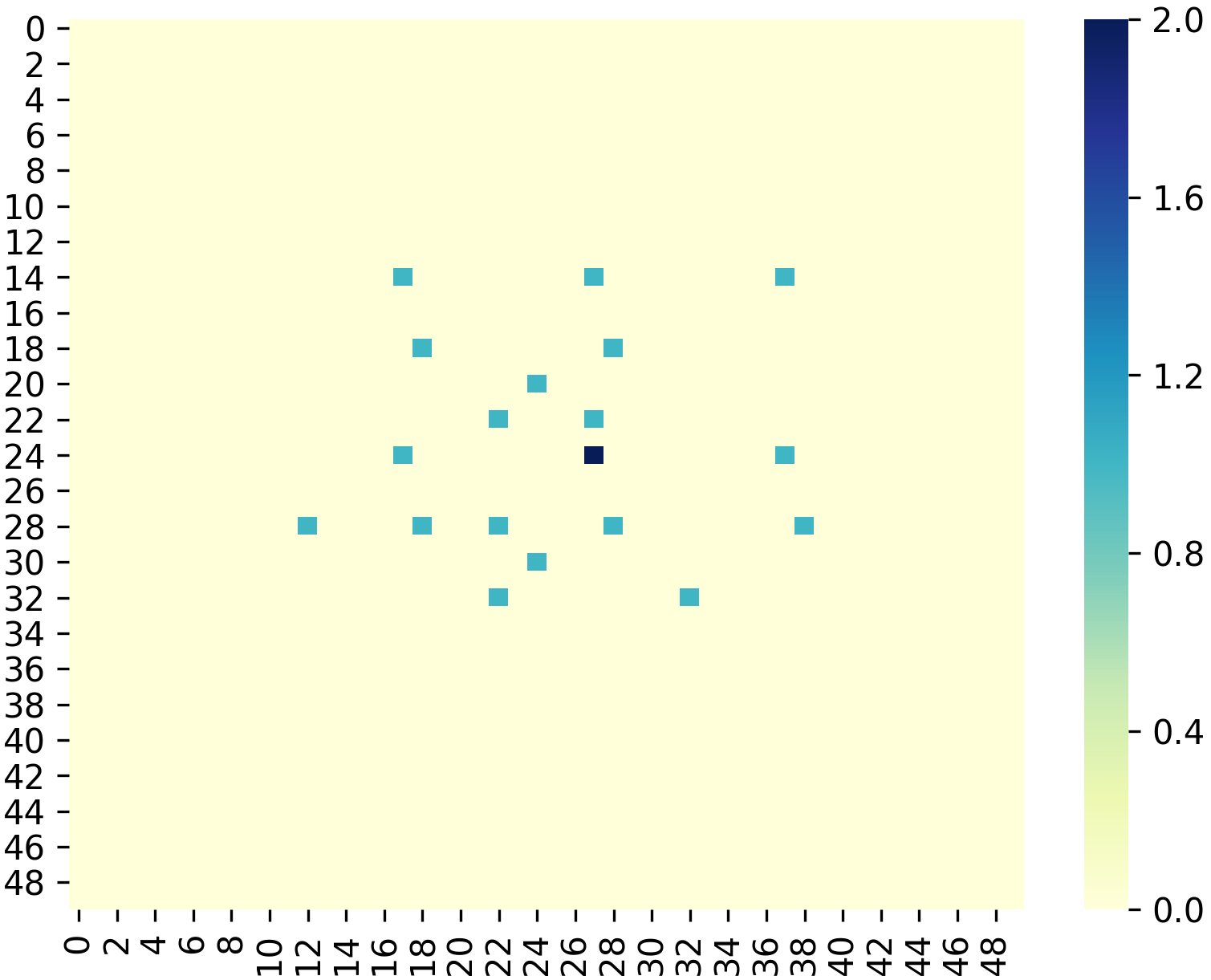}}}
	
	\vspace{-0.3cm}
	
	\subfigure[transportation]{\label{fig:channel14}\includegraphics[width=0.24\linewidth]{{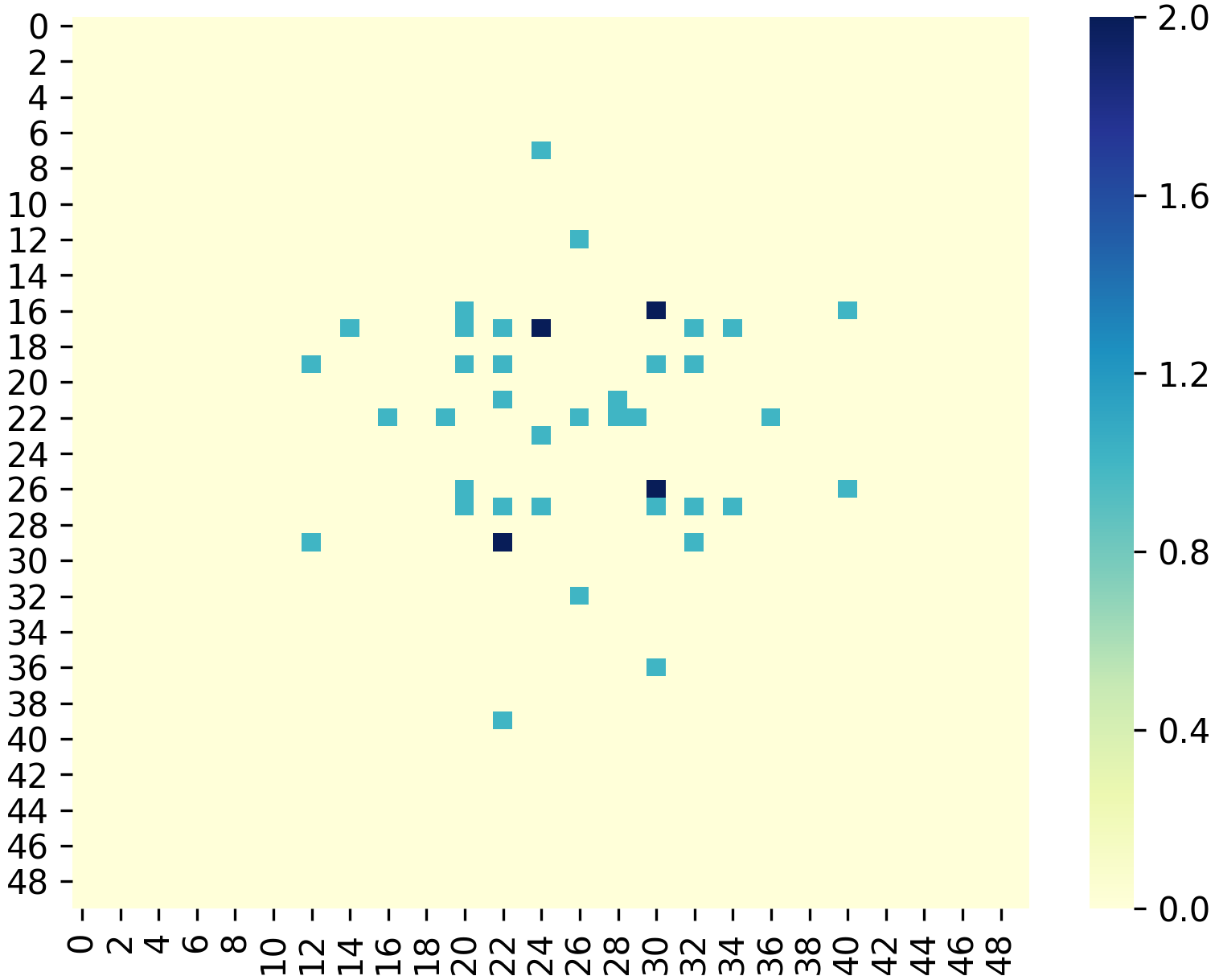}}}
	\subfigure[specific address]{\label{fig:channel18}\includegraphics[width=0.24\linewidth]{{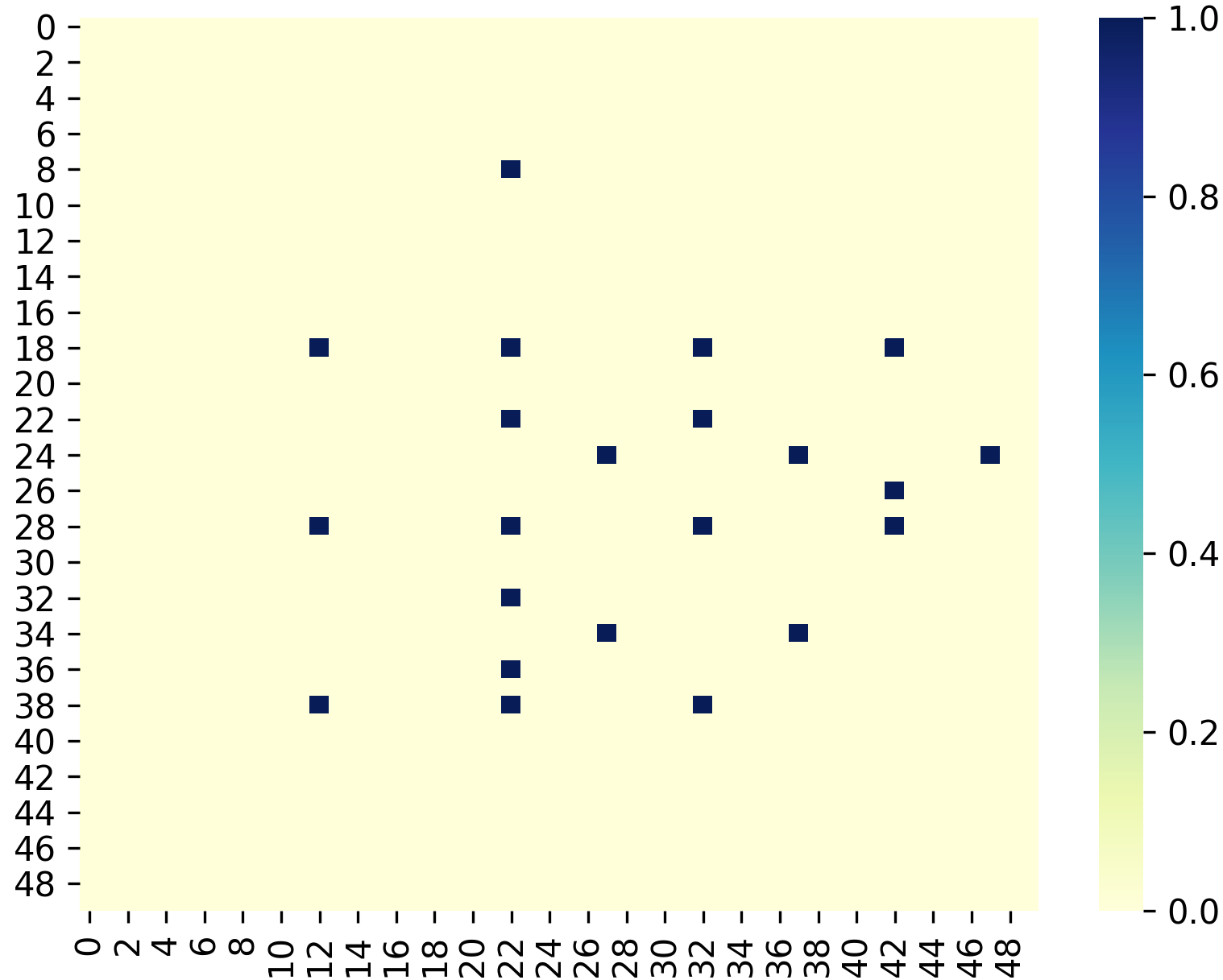}}}
	\subfigure[finance]{\label{fig:channel15}\includegraphics[width=0.24\linewidth]{{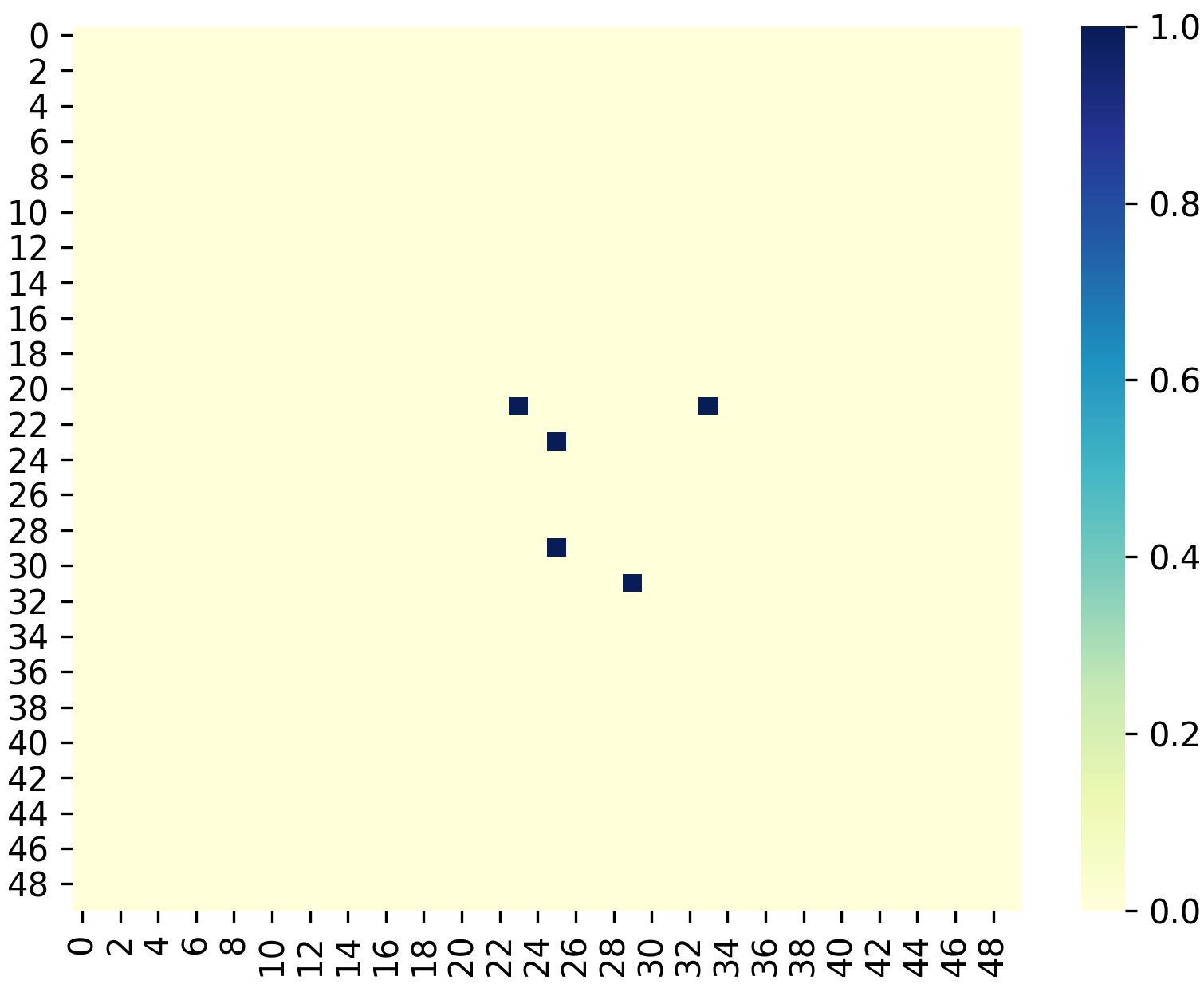}}}
	\subfigure[company]{\label{fig:channel16}\includegraphics[width=0.24\linewidth]{{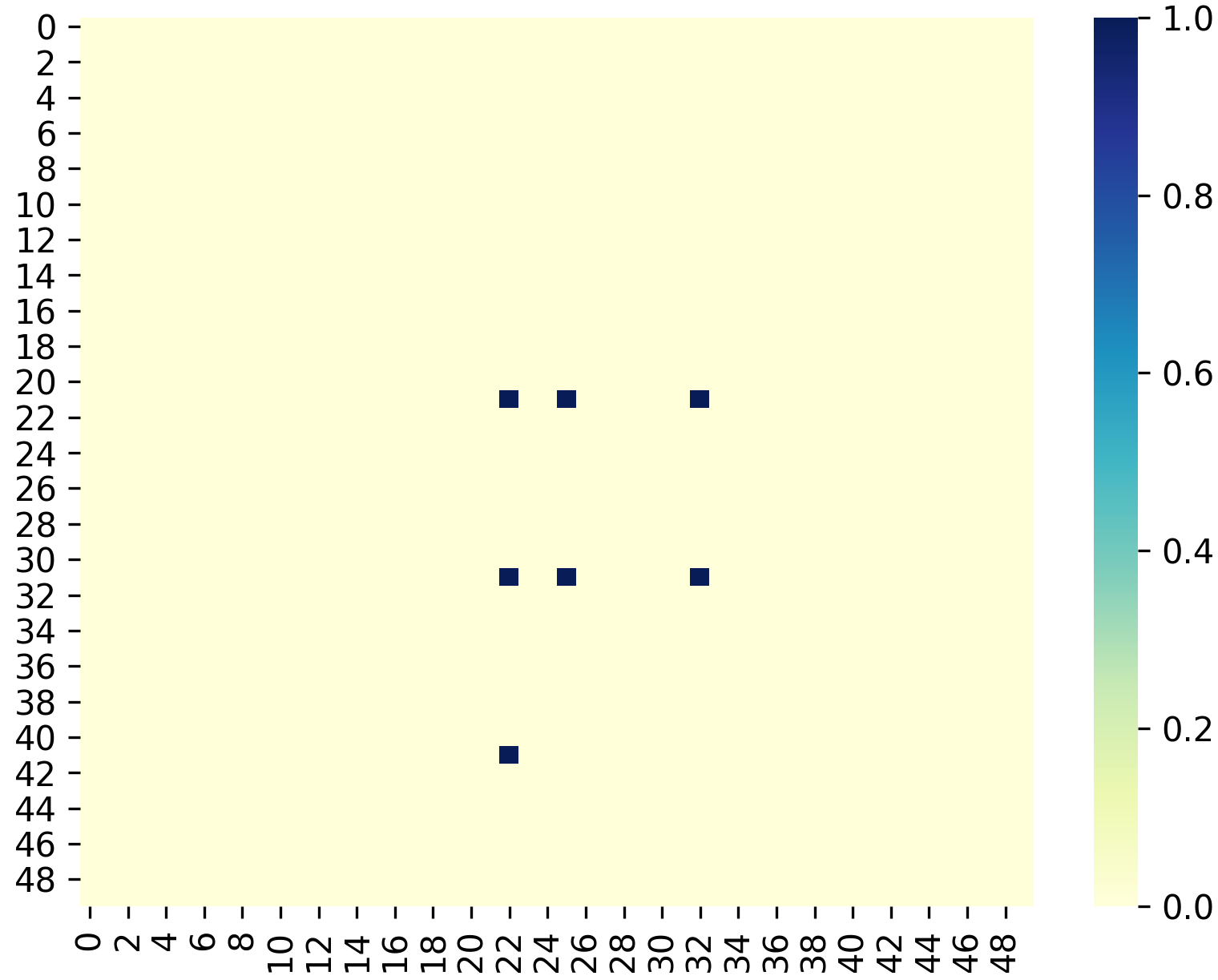}}}
	
	\vspace{-0.1cm}
	
	\caption{Visualization for different POI categories of one generated configuration.}
	\label{fig:visual_solution}
	\vspace{-0.8cm}
\end{figure*}

We quantify a land-use configuration as a longitude-latitude-channel tensor.
So, what is the generated situation for each channel (POI category)?
To check it, we visualize the POI distribution of each channel.
The visualization results are shown in  Figure \ref{fig:visual_solution}, in which the darker color of the block indicates the number of POI in the corresponding block is larger.
An interesting observation standing out is that the POI distributions of different categories show their unique patterns.
For example, transportation pots are more concentrated, while food service related POIs are more dispersed across the area;
the distribution of car service spots is very similar to the recreation service, and the possible reason is that recreation service spots may occupy many parking lots which potentially attract car services.
The observation shows that LUCGAN$^+$ is capable of capturing characteristics of POI distribution of different categories at the same time.
From another perspective, the observation also reflects that LUCGAN$^+$ is able to capture the mutual interactions and constraints among different kinds of POIs. 
Thus, LUCGAN$^+$ is superior and effective for generating land-use configurations automatically.

\section{Related Work}

\textbf{Spatio-temporal Data Mining}
Spatio-temporal data mining refers to the process of discovering the pattern and knowledge from the data related to space and time ~\cite{atluri2018spatio}. 
Owing to the spatio-temporal data is closely relevant to our real life, many researchers attempted to extract the patterns hidden behind the data for improving the urban life quality ~\cite{wang2018deep,wang2021automated,wang2021reinforced,wang2021towards,zhou2020deep,liu2018modeling}. 
For instance, Wang et al. employed deep learning approaches to forecast the travel demand of individuals based on the travel order data collected by car-hailing company ~\cite{8489530}.
Zhao et al. predicted the air quality index by considering spatio-temporal relatedness ~\cite{zhao2017incorporating}. 
Wang et al. employed reinforcement learning and spatial knowledge graph to conduct mobile user profiling ~\cite{wang2020incremental}.
Yuan et al. utilized topic model to discover urban functional zone based on POI data and taxi trajectories data ~\cite{yuan2014discovering}.
Wang et al. used peer and temporal-aware representation learning to analyze the driving behavior based on GPS trajectory data ~\cite{wang2018you}.
Liu et al. studied the mobility patterns of traffic flows for bus routing optimization ~\cite{liu2017intelligent}.
Du et al. provided a systematic study to capture the spatio-temporal dynamics of passenger transfers for crowdedness-aware route recommendations ~\cite{du2018smarttransfer}.
% Thus, the spatio-temporal data mining is significant and meaningful.
In this paper, to reduce the heavy workload of urban planners and accelerate the urban planning process, we expect to utilize spatio-temporal data for the urban planning pattern discovering.

\textbf{Representation Learning.} 
The objective of representation learning is to preserve the information of original data into a low-dimensional feature space.
In general, there are three types of representation learning models:
(1) probabilistic graphical models;
(2) manifold learning models;
(3) auto-encoder models.
The probabilistic graphical models build a complex Bayesian network system to learn the representation of uncertain knowledge buried in original data \cite{qiang2019learning}.
% But it is hard to find the topology structure of the Bayesian network and calculate the transfer probability among nodes in the graphical model.
The manifold learning models infer low-dimensional manifold of original data based on neighborhood information by non-parametric approaches \cite{zhu2018image}.
% The models own a solid theoretical basis, but the resolution process needs much time.
The auto-encoder models learn the latent representation by minimizing the reconstruction loss between original and reconstructed data \cite{otto2019linearly}.
% In this paper, we utilize the auto-encoder paradigm to learn the representation of the surrounding contexts.
In the spatio-temporal data mining domain, to capture the characteristics of spatial entity (i.e. city, geographical area), representation learning achieves great success \cite{fu2018representing,wang2018learning,wang2020exploiting,chandra2019collective,fu2019efficient}.
% ,zhang2019unifying
For instance, to analyze the individual driving behaviors,  Wang et al. utilized representation learning to mine the spatio-temporal characteristics of GPS trajectory data. \cite{wang2019spatiotemporal}.
Du et al. proposed a new spatial representation learning framework to capture the static and dynamic characteristics among the spatial entities for predicting housing price
\cite{8970913}. 
Wang et al. employed a spatio-temporal representation learning module to extract the features of cyber attack in a graph for cyber attack detection ~\cite{9338292}. 
In this paper, to incorporate the surrounding context characteristics into our framework, we employ representation learning to preserve the spatial attributed graphs constructed by the contexts into low-dimensional vectors.

\textbf{Generative Adversarial Networks.}
Recently, Generative Adversarial Networks (GAN) attract tremendous attention of researchers \cite{zhang2020curb,zhang2019trafficgan}.
GAN algorithms can be classified into three categories from the task-driven perspective.
(1) Semi-supervised learning GANs. 
Usually, a complete labeled data set is difficult to obtain, and the semi-supervised learning GANs can utilize unlabeled data or partially labeled data to train an excellent classifier \cite{ding2018semi,liu2020catgan}. 
For instance, Akcay et al. designed a semi-supervised GAN anomaly detection framework that achieved good performance \cite{akcay2018ganomaly}.
(2) Transfer learning GANs.
Many researchers utilize the transfer learning GANs to transfer knowledge among different domains \cite{hoffman2018cycada,tzeng2017adversarial}.
For instance, Choi et al. built an unified GAN to translate the images among different style fields \cite{choi2018stargan}.
(3) Reinforcement learning GANs.
Reinforcement learning (RL) is incorporated into GANs to improve the generative performance \cite{sarmad2019rl}.
For instance, Ganin et al. combined reinforce learning and GAN to synthesize high-resolution images \cite{ganin2018synthesizing}.
Aforementioned works indicate that GANs are capable of capturing the characteristics of the original data distribution and generate new data samples based on the distribution.
Such observation motivates us to utilize the learning paradigm of GANs as the main framework of our automatic urban planner.

\textbf{Urban Planning.}
Urban planning is a complex and interdisciplinary research domain \cite{adams1994urban}.
Urban experts  need to consider lots of factors such as government policy, environmental protection, and more for designing appropriate land-use configurations \cite{niemela1999ecology,adams1994urban,wang2021deep,oliveira2010evaluation}.
Meanwhile, different areas have various planning goals.
For example, Barton et al. focused on constructing an urban planning solution for human health and well-being \cite{barton2013healthy}.
John et al. discussed the relationship between urban planning and real estate development \cite{ratcliffe2004urban}.
Indeed, it is difficult to generate a good urban planning solution objectively.
Recently, with the development of artificial intelligence (AI), many researchers focus on making the process of urban planning become smart and automated.
These methods always build up a GAN model to generate the layout of a space based on the realistic architecture or designing image ~\cite{li2018layoutgan,nauata2020house,noyman2020deep}.
For instance, Albert et al. utilized generative adversarial networks to generate the complex and spatial organizations observed in global urban patterns based on footprint data ~\cite{albert2018modeling}.
Bachl et al. proposed a new conditional GAN framework to learn the architecture features of major cities for generating the image of buildings which do not exist before ~\cite{bachl2019city}.
% Tian et al. utilized Pix2Pix2HD model to generate the site layout based on the open source geographic information system data ~\cite{tian2020suggestive}.
These works have had a lot of success, but they have one drawback: they require expert layout data in order to train AI models.
In addition, some researchers use transfer learning to transfer spatial knowledge across many cities to increase the generalization of spatial AI models and learning efficiency ~\cite{jiang2021transfer,10.1145/3366423.3380210}.
These works are capable of perceiving human mobility patterns, which gives a decent foundation for urban planning, but they are unable to immediately develop successful urban layouts.
Compared to these works, our framework LUCGAN$^+$ has no strict condition for data collection. 
We focus on producing customized land-use configurations based on geographical spatial data such as POI, traffic data, economic data, demographic data, etc.
These data resources are always publicly available, which makes our framework have good flexibility and generalization.

\section{Conclusion Remarks}
In order to generate a suitable and excellent land-use configuration solution objectively and reduce the heavy burden of urban planning specialists, we propose an automatic land-use configuration planner framework.
This framework generates the land-use configuration based on the surrounding contexts.
Specifically, we first collect a set of land-use configurations and corresponding surrounding contexts.
Then, we construct spatial attributed graphs that contain explicit features such as value-added space, POI distribution, traffic conditions, and more of surrounding contexts, and preserve the information of the graphs into the surrounding embeddings.
Next, we employ our proposed automatic urban planner model to generate well-planned land-use configurations based on the embeddings.
Finally, through extensive experiments, we find that LUCGAN$^+$ is more effective and robust than other baseline models.
In addition, different square sizes affect the generative ability of LUCGAN$^+$, so users should adopt suitable segmentation scheme for land-use configurations based on their requirements. 
Moreover, LUCGAN$^+$ is capable of customizing land-use configurations based on hyperparameter $Q$.
Furthermore, LUCGAN$^+$ not only generates the POI distribution of whole area but also provides the generation of each POI category.

\bibliographystyle{ACM-Reference-Format}
\bibliography{acmart}

\end{document}